\documentclass[lettersize,journal]{IEEEtran}
\usepackage{amsmath,amsfonts,amsthm,amssymb}
\usepackage{algorithmic}
\usepackage{algorithm}
\usepackage{array}
\usepackage{textcomp}
\usepackage{stfloats}
\usepackage{url}
\usepackage{verbatim}
\usepackage{graphicx}
\usepackage{cite}
\usepackage{bm}
\usepackage{hyperref}
\usepackage{cleveref}
\usepackage{subfigure}
\usepackage{multirow}

\newtheorem{definition}{Definition}
\newtheorem{theorem}{Theorem}

\newtheorem{lemma}{Lemma}
\newtheorem{assumption}{Assumption}
\newtheorem{corollary}{Corollary}

\newcommand{\bx}{\bm{x}}
\newcommand{\bU}{\bm{U}}
\newcommand{\bz}{\bm{z}}
\newcommand{\bA}{\bm{A}}
\newcommand{\bu}{\bm{u}}
\newcommand{\by}{\bm{y}}
\newcommand{\bb}{\bm{b}}
\newcommand{\bX}{\bm{X}}
\newcommand{\blam}{\bm{\lambda}}

\begin{document}

\title{Low Tensor-Rank Adaptation of Kolmogorov--Arnold Networks}

\author{Yihang Gao\thanks{Department of Mathematics, National University of Singapore}, Michael K. Ng\thanks{Department of Mathematics, Hong Kong Baptist University}, 
Vincent Y. F. Tan\thanks{Department of Mathematics and Department of Electrical and Computer Engineering, National University of Singapore}
}



\maketitle

\begin{abstract}
Kolmogorov--Arnold networks (KANs) have demonstrated their potential as an alternative to multi-layer perceptions (MLPs) in various domains, especially for science-related tasks. However, transfer learning of KANs remains a relatively unexplored area. In this paper, inspired by Tucker decomposition of tensors and evidence on the low tensor-rank structure in KAN parameter updates, we develop low tensor-rank adaptation (LoTRA) for fine-tuning KANs.
We study the expressiveness of LoTRA based on Tucker decomposition approximations. Furthermore, we provide a theoretical analysis to select the learning rates for each LoTRA component to enable efficient training. Our analysis also shows that using identical learning rates across all components leads to inefficient training, highlighting the need for an adaptive learning rate strategy.
Beyond theoretical insights, we explore the application of LoTRA for efficiently solving various partial differential equations (PDEs) by fine-tuning KANs. Additionally, we propose Slim KANs that incorporate the inherent low-tensor-rank properties of KAN parameter tensors to reduce model size while maintaining superior performance.
Experimental results validate the efficacy of the proposed learning rate selection strategy and demonstrate the effectiveness of LoTRA for transfer learning of KANs in solving PDEs. Further evaluations on Slim KANs for function representation and image classification tasks highlight the expressiveness of LoTRA and the potential for parameter reduction through low tensor-rank decomposition.
\end{abstract}


\section{Introduction}
The Kolmogorov--Arnold representation theorem (KART) states that any multivariate continuous function $f: \mathbb{R}^{n} \to \mathbb{R}$ can be decomposed into a sum of univariate functions. Specifically, there exists a set of univariate functions $\{\phi_{p,q}\}_{p=1,q=1}^{n, 2n+1}$ and $\{\Phi_{q}\}_{q=1}^{2n+1}$ such that
\begin{equation}
\label{eq_KART}
    f(\bm{x}) = \sum_{q=1}^{2n+1} \Phi_{q}\left(\sum_{p=1}^{n} \phi_{p,q}\left( x_{p}\right)\right),
\end{equation}
where $\bx = [x_1, x_2, \ldots, x_n]^{\top}$. 
Inspired by this theorem, researchers have explored novel network architectures, as an alternative to multi-layer perceptrons (MLPs) guaranteed by the Universal Approximability Theory (UAT). 
However, earlier network designs based on KART have been underwhelming, mainly because they are inherently limited to two layers and a width restricted to at most twice the input dimension, as prescribed by \Cref{eq_KART}. 
These models typically represent the implicit univariate functions  $\{\phi_{p,q}\}_{p=1,q=1}^{n, 2n+1}$ and $\{\Phi_{q}\}_{q=1}^{2n+1}$ using polynomials or splines. Although theoretically sound, such models often fail to perform competitively in practical applications, lagging behind MLPs.
Recently, Liu et al.~\cite{liu2025kan} proposed a KART-based model known as Kolmogorov--Arnold Networks (KANs), which extends the architecture to support multiple layers and arbitrary width. 
This design overcomes the stringent limitations on width and depth inherent in traditional KART-based networks, enabling greater expressiveness and flexibility.

KANs have demonstrated remarkable empirical performance across a wide range of domains, including computer vision, time series forecasting, reinforcement learning, physics-informed machine learning, and large language models. 
Li et al. \cite{li2024u} developed U-KAN, which replaces the bottleneck layers of U-Net with KAN layers, achieving higher accuracy with reduced computational cost in medical image segmentation and generation tasks. This improvement is attributed to the potentially higher expressiveness of KANs compared to MLPs. 
Kich et al. \cite{kich2024kolmogorov} explored the application of KANs as function approximators in proximal policy optimization for reinforcement learning, demonstrating that KANs can match the performance of MLPs while requiring fewer parameters.
Notably, KANs excel in symbolic and function representation tasks. In physics-informed machine learning, KANs consistently match or outperform MLPs, especially in approximating solutions of partial differential equations (PDEs), as reported by Wang et al.~\cite{wang2025physics}. 
To model the separable properties inherent in some PDE solutions, a separable physics-informed KAN architecture was proposed~\cite{jacob2024spikans}. This design processes each block independently using individual KAN models before synthesizing them in the final stage, significantly reducing model complexity by incorporating the natural separable structure of PDE solutions.
More interesting applications and investigations of KANs can be found in \cite{shukla2024a,hou2024comprehensive,somvanshi2024survey,firsov2024hyperkan,zhou2024kan,wang2024expressiveness,mehrabian2024implicit,howard2024finite,rigas2024adaptive,kundu2024kanqas,bodner2024convolutional,drokin2024kolmogorov}.

As illustrated above, one of the most significant advantages of KANs over MLPs lies in their superior performance in science-related tasks, particularly in physics-informed machine learning. 
Here, we mainly focus on using KANs to approximate solutions of PDEs. However, in practice, even slight variations in physical parameters or conditions can result in changes to the solutions.
Instead of solving a single PDE and storing the corresponding network, it is often necessary to solve a class of PDEs with varying physical parameters. This setting requires solving PDEs multiple times and storing numerous networks, which significantly increases computational costs and storage requirements.

However, the transfer learning of KANs has not been investigated, despite its potential importance in enhancing the efficiency of model training and storage, particularly in physics-informed machine learning applications.
Inspired by the recently popular low-rank adaptation (LoRA) technique used in Transformer models, which significantly accelerates the fine-tuning process for large language models, and the evidence on low tensor-rank structure of parameter updates of KANs, we introduce the low tensor-rank adaptation (LoTRA) for the tensor parameters of KANs.
In this approach, we first pre-train the KAN model on a given task with full tensor parameters (denoted as $\mathcal{A}$) being updated. For a new task, instead of updating the entire tensor parameter $\mathcal{A}$, we apply the Tucker decomposition to the tensor, adapting it as $\mathcal{A} + \mathcal{G} \times_{1} \bU^{(1)} \times_{2} \bU^{(2)} \times_{3} \bU^{(3)}$. Here, $\{\mathcal{G},\bU^{(1)},\bU^{(2)},\bU^{(3)}\}$ are trainable parameters, providing a low tensor-rank adaptation for the tensor parameters of KANs.
In this adaptation, $\mathcal{G}$ is the core tensor, and $\{\bU^{(1)},\bU^{(2)},\bU^{(3)}\}$ are transformation matrices. 
If the core tensor $\mathcal{G}$ is significantly smaller than  $\mathcal{A}$, then the total parameter size of $\{\mathcal{G},\bU^{(1)},\bU^{(2)},\bU^{(3)}\}$ is much smaller than that of $\mathcal{A}$, resulting in efficient fine-tuning and reduced storage. 
In this framework, the original tensor $\mathcal{A}$ captures and retains the shared information across tasks from the pre-training stage, while $\{\mathcal{G},\bU^{(1)},\bU^{(2)},\bU^{(3)}\}$ further adapt the model to task-specific characteristics in new tasks. This is the first work studying the transfer learning of KANs through low tensor rank adaptation, paving the way for more efficient and scalable applications.  Our contributions are summarized as follows:

\noindent (i) We introduce the low tensor-rank adaptation (LoTRA) for the transfer learning of KANs, using Tucker decomposition to achieve efficient and effective adaptation.

\noindent (ii) We theoretically show the expressiveness of LoTRA and analyze the efficient training of KANs with LoTRA. Our theoretical results offer a learning rate selection strategy to fine-tune KANs with LoTRA efficiently.

\noindent (iii) The proposed LoTRA method adapts well to physics-informed machine learning with KANs, enabling efficient training and significantly reduced storage for solving a class of PDEs. 
Additionally, the LoTRA framework motivates the development of Slim KANs, which adopt a low tensor-rank structure in parameter tensors to reduce model size while preserving performance.

\noindent (iv) We conduct comprehensive experiments on solving a class of PDEs with varying parameters and solutions, validating the developed learning rate selection strategy and showing the effectiveness of LoTRA in fine-tuning KANs. Additional experiments on Slim KANs further demonstrate the expressiveness of LoTRA and the potential of integrating parameter-efficient models into advanced architectures for broader applications.

\section{Background and Preliminaries}

In this section, we first introduce the notation used throughout the paper. We then review the mathematical definition and key concepts of KANs. Finally, we discuss the mathematical formulation of Tucker decomposition for tensors, which serves as a foundation for the methodology section.

\subsection{Notation}
In this paper, we use bold lowercase letters (e.g., $\bm{x}$), bold capital letters (e.g., $\bm{A}$), and calligraphic uppercase letters (e.g., $\mathcal{A}$) to denote vectors, matrices and tensors, respectively. Scalars are represented using regular (non-bold) letters (e.g., $a$). 
The reshape operation, denoted as $\text{reshape}(\mathcal{A};i;a,b)$, restructures the tensor $\mathcal{A}$ along mode $i$, transforming it into an $a \times b$ matrix.
We use $[L]$ to denote the set $\{1,2,\ldots,L\}$ for positive integers $L$.

\subsection{Kolmogorov--Arnold Networks}
As defined in~\cite{liu2025kan}, the $(\ell+1)$-st layer $\bm{z}_{\ell+1}=[z_{\ell+1,1},\ldots,z_{\ell+1,n_{\ell+1}}]^{\top}$ of a KAN admits
\begin{equation}
\label{eq_original_kan}
    z_{\ell+1,q} = \sum_{p=1}^{n_{\ell}} \phi_{\ell,p,q}(z_{\ell,p}), \quad q \in [n_{\ell+1}],
\end{equation}
where $n_{\ell}$ is the width of the $\ell$-th layer and $\phi_{\ell,p,q}$ denotes the $(p,q)$-th representation function in the $\ell$-th layer.
This formulation is consistent with the KART presented in \Cref{eq_KART}, but it introduces some key differences.
Notably, KANs do not impose restrictions on width ($n_{\ell}$ can exceed twice the input dimension $n$) or depth (e.g., $\ell$ can be much greater than $2$).
However, the implicit representation functions $\{\phi_{\ell,p,q}\}_{\ell,p,q}$ in each layer are unknown, and must be approximated using practical function classes.
In this sense, the development of KANs draws inspiration both from the theoretical principles of KART and the flexibility of MLPs.

In practice, the implicit representation functions are expressed by combinations of basis functions, i.e.,
\begin{equation*}
    \phi_{\ell,p,q}(z) = \sum_{k=1}^{n_d} a_{\ell,p,q,k} b_{k}(z),
\end{equation*}
where $\{b_{k} \}_{  k \in [n_d]}$ is a set of basis functions and $\mathcal{A}^{\ell} := (a_{\ell,p,q,k}) \in \mathbb{R}^{n_{\ell} \times n_{\ell+1} \times n_{d}}$ denotes the parameter tensor at $\ell$-th layer. 
The basis functions adopted can include Chebyshev polynomials~\cite{ss2024chebyshev}, Legendre polynomials~\cite{anonymous2025legendrekan}, Fourier series~\cite{xu2024fourierkan}, wavelet functions~\cite{bozorgasl2024wav}, Bernoulli polynomials, Fibonacci polynomials, Jacobi polynomials~\cite{seydi2024exploring}, B-splines~\cite{liu2025kan}, as well as rational and fractional polynomials~\cite{aghaei2024rkan,aghaei2025fkan}, due to the universal approximability of those basis functions.

In summary, the transformation of KANs from $\ell$-th layer to $(\ell+1)$-th layer is formulated as
\begin{equation}
\label{eq_kan}
    z_{\ell+1,q} = \sum_{p=1}^{n_{\ell}} 
    \sum_{k=1}^{n_d} a_{\ell,p,q,k} b_{k}(z_{\ell,p}), \quad q \in [n_{\ell+1}],
\end{equation}
where $\bz_{\ell} = (z_{\ell,1},z_{\ell,2},\ldots,z_{\ell,n_{\ell}})^{\top} \in \mathbb{R}^{n_{\ell}}$ and $\bz_{\ell+1} = (z_{\ell+1,1},z_{\ell+1,2},\ldots,z_{\ell+1,n_{\ell+1}})^{\top} \in \mathbb{R}^{n_{\ell+1}}$ denote the neurons at the $\ell$-th and $(\ell+1)$-th layers, respectively, and $\mathcal{A}^{\ell} := (a_{\ell,p,q,k}) \in \mathbb{R}^{n_{\ell} \times n_{\ell+1} \times n_{d}}$ represents the parameter tensor at $\ell$-th layer.

\subsection{Tucker Decomposition}

Tensor decomposition methods have shown significant success in various domains, such as hyperspectral image processing~\cite{zhuang2021hyperspectral,xue2019nonlocal,yokota2016smooth}, multidimensional time series analysis~\cite{chen2021bayesian,arulampalam2002tutorial}, and high-dimensional machine learning~\cite{sidiropoulos2017tensor,shen2017tensor}. 
Tucker decomposition method~\cite{tucker1963implications} compresses the tensor into a smaller core tensor with transformation matrices applied along each mode, whose structure is similar to the singular value decomposition (SVD) of matrices. Other tensor decomposition methods include CANDECOMP/PARAFAC (CP) method~\cite{kiers2000towards}, tensor SVD~\cite{song2020robust,kilmer2011factorization}, CANDECOMP with linear constraints (CANDELINC)~\cite{carroll1980candelinc}, and parallel factors for cross products (PARAFAC2)~\cite{harshman1972parafac2}, among others.

In this paper, we make use of the Tucker decomposition of tensors. As shown in \Cref{eq_kan}, the parameters of KANs for each layer are represented by a third-order tensor, whereas the parameters of MLPs are matrices. Tucker decomposition is considered a generalization of matrix SVD to tensors. For a given third-order tensor $\mathcal{A} \in \mathbb{R}^{n_1 \times n_2 \times n_3}$, Tucker decomposition expresses $\mathcal{A}$ as a core tensor $\mathcal{G} \in \mathbb{R}^{r_1 \times r_2 \times r_3}$, and three transformation matrices applied to each mode, $\bU^{(1)} \in \mathbb{R}^{n_1 \times r_1}$, $\bU^{(2)} \in \mathbb{R}^{n_2 \times r_2}$ and $\bU^{(3)} \in \mathbb{R}^{n_3 \times r_3}$, such that 
\begin{equation}
    \mathcal{A} = \mathcal{G} \times_{1} \bU^{(1)} \times_{2} \bU^{(2)} \times_{3} \bU^{(3)},
\end{equation}
where $\times_{i}$ denotes the mode-$i$ product of the core tensor $\mathcal{G}$ with the transformation matrices $\bU^{(i)}$ (for $i = 1,2,3$). If the tensor $\mathcal{A}$ has a low Tucker rank, then $r_{i} < n_{i}$ (for $i=1,2,3$). This decomposition closely parallels matrix SVD, where the core tensor $\mathcal{G}$ plays a role analogous to the singular values, while $\bU^{(1)}$, $\bU^{(2)}$ and $\bU^{(3)}$ serve as basis matrices. The essential information of the tensor is compressed into the core tensor $\mathcal{G}$ and can be fully reconstructed using the appropriate transformation matrices. If the tensor $\mathcal{A}$ has an extremely low Tucker rank (e.g., $r_i \ll n_i$, for $i=1,2,3$), then the parameter size of $\{\mathcal{G}, \bU^{(1)}, \bU^{(2)}, \bU^{(3)}\}$ is much smaller than that of the original tensor, leading to the effective compression by Tucker decomposition.

The higher-order singular value decomposition (HOSVD) method~\cite{de2000multilinear} provides a numerical approach to determine the Tucker decomposition for tensors. 
Let $\bm{A}^{(i)}$ denote the mode-$i$ unfolding  (or matricization) of the tensor $\mathcal{A}$,   defined as
\begin{equation*}
    \bA^{(i)} = \text{reshape}\left(\mathcal{A};i;n_{i},\frac{n_1 n_2 n_3}{n_i} \right).
\end{equation*}
In this unfolding, the mode-$i$ fibers of $\mathcal{A}$ are arranged as columns in  $\bA^{(i)}$. This operation reorganizes the tensor by flattening all dimensions except the $i$-th mode. The transformation matrices $\{\bU^{(1)},\bU^{(2)},\bU^{(3)}\}$ are then obtained by performing SVDs on all mode-$i$ unfolding matrices $\bA^{(1)}$, $\bA^{(2)}$, and $\bA^{(3)}$. 
If any of the unfolding matrices contain zero singular values, the core tensor $\mathcal{G}$ has a smaller size than the original tensor $\mathcal{A}$, indicating the presence of a low-rank structure in $\mathcal{A}$.

\section{Methods}
In this section, we first provide evidence supporting the potential low tensor-rank structure of the parameter updates, which motivates the development of the low tensor-rank adaptation (LoTRA) for transfer learning of KANs. 
We then theoretically analyze the approximation and expressiveness capabilities of LoTRA, using well-established results on Tucker decomposition.
Furthermore, we propose a theoretically guided strategy for learning rate selection of each component of LoTRA to achieve efficient training using gradient descent. In contrast, we demonstrate theoretically that using the same learning rate scale for all components is inefficient for LoTRA.

\subsection{Motivation and Evidence}
\label{sec_motivation}
The concept of low-rank adaptation is widely recognized in transfer learning and domain adaptation.
In multi-task learning, the parameter matrix is typically decomposed into global patterns shared across tasks and task-specific adaptations within the shared subspace~\cite{caruana1997multitask,zhang2021survey}. This decomposition captures shared information while regularizing models to prevent overfitting. Similarly, in domain adaptation, both source and target domains are projected into a shared feature space through low-rank mappings, reflecting the existence of underlying shared low-rank subspaces~\cite{mansour2008domain,daume2007frustratingly}.
In the fine-tuning of large language models, Hu et al.~\cite{hu2022lora} empirically demonstrated that weight updates during adaptation exhibit low intrinsic matrix rank. This observation led to the development of low-rank adaptation (LoRA) for Transformer models. 

Based on their observations, we have reason to hypothesize that the update of KANs parameters has a low tensor-rank structure in transferring learning tasks, specifically, fine-tuning on new tasks based on the pre-trained models. The potential intrinsic low tensor-rank property has been  seen in  some empirical investigations on KANs. 
For example, function mappings in science domains usually involve specific structures that may lead to similarities between implicit representation functions $\{\phi_{\ell,p,q}\}$ in \Cref{eq_original_kan}. Specifically, the symbolic representation of the function $f\left(x_1, x_2, x_3, x_4\right)=\exp \left(\frac{1}{2}\left(\sin \left(\pi\left(x_1^2+x_2^2\right)\right)+\sin \left(\pi\left(x_3^2+x_4^2\right)\right)\right)\right)$ is discussed in Liu et al.~\cite{liu2025kan}, where a three-layer KAN fits the function well. However, the first layer involves mostly the square operation applied to all coordinates and the second layer only represents  the $\sin$ function. The final layer produces the outputs after passing through the exponential function. This example shows the repeated operations in each layer, implying the possibility of compression of parameter tensors of KANs. 
Similar properties have been observed in the solutions of PDEs studied in physics-informed KANs, as shown in Shukla et al.\cite{shukla2024a} and Wang et al.\cite{wang2025physics}, further supporting the possibility of low-rank structures in KAN parameter tensors, particularly in symbolic representation and physics-informed machine learning.

\begin{figure}[ht]
    \centering
    \subfigure[Singular values for the pre-trained model parameters]{%
        \includegraphics[width=0.23\textwidth]{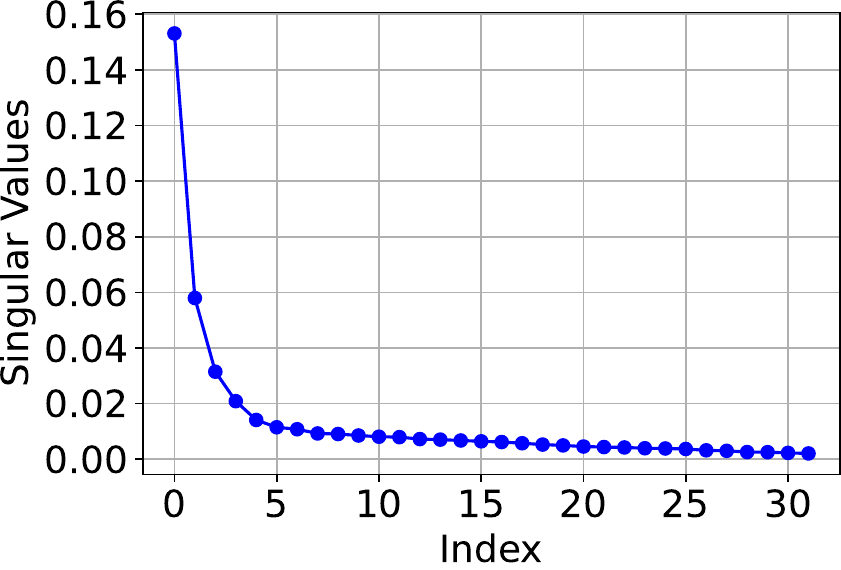} \label{fig:evidene_fig1}}
    \subfigure[Singular values for fine-tuned updates]{%
        \includegraphics[width=0.23\textwidth]{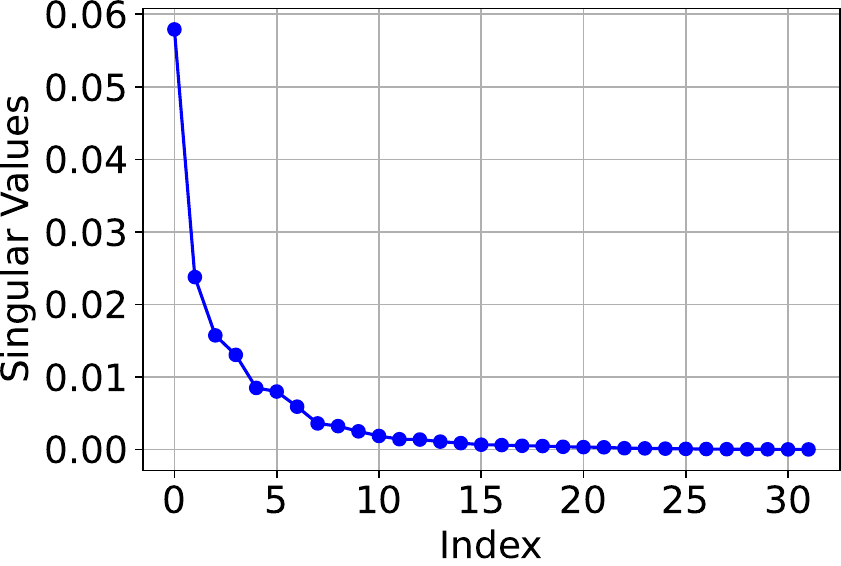} \label{fig:evidene_fig2}}
    \caption{Singular values for the pre-trained model parameters and fine-tuned updates, demonstrating the low Tucker rank structure.}
    \label{fig:evidence}
\end{figure}

To further validate our hypothesis regarding the low-rank updates of KAN parameter tensors, we conduct a simple transfer learning experiment on function representation. We use a three-layer KAN with a hidden dimensions $n_{1} = n_{2} = 32$ and the number of basis functions $n_d = 32$ to represent and pre-train on the function $u(x_1, x_2)=\sin \big(\frac{\pi}{2}\big(1-\sqrt{x_1^2+x_2^2}\big)^{2.5}\big)$. We then fine-tune the KAN model on a new task with the modified objective function: $u(x_1, x_2)=\sin \big(\frac{\pi}{2}\big(1-\sqrt{x_1^2+x_2^2}\big)^{2.5}\big) + \sin \big(\frac{\pi}{2}\big(1-\sqrt{x_1^2+x_2^2}\big)\big)$. 
Our goal is to empirically investigate whether the parameter tensor updates in the fine-tuned model exhibit a low tensor-rank structure. We measure the tensor rank using the Tucker rank and determine it using HOSVD, where the singular values of the mode-unfolded matrices correspond to the Tucker rank of the tensors. Specifically, zero singular values in the unfolded matrices indicate a low Tucker rank.
Our observations reveal that KANs at the pre-training stage already exhibit a low Tucker rank structure, where  most of the singular values are  small, as visualized in \Cref{fig:evidene_fig1}. Furthermore, we compute the parameter tensor updates during fine-tuning relative to the pre-trained model and parameters. As shown in \Cref{fig:evidene_fig2}, the singular values confirm the low Tucker rank structure in the parameter tensor updates.

\subsection{Low Tensor-Rank Adaptation}
Building on observations of low-rank structures in previous studies across various domains, as well as the empirical investigations of a simple transfer learning example, we develop an efficient and effective transfer learning method for fine-tuning KANs on new tasks, given a pre-trained model. 
The developed low tensor-rank adaptation (LoTRA) method preserves shared information across tasks and adapts to new tasks with integrated low-rank structures on parameter tensor updates. 
For simplicity, we express \Cref{eq_kan} as 
\begin{equation*}
    \bm{z}_{\ell + 1}:= \Phi_{\ell}\left(\bm{z}_{\ell};\mathcal{A}_{\ell}\right),
\end{equation*}
where $\mathcal{A}_{\ell} \in \mathbb{R}^{n_{\ell} \times n_{\ell + 1} \times n_{d}}$ is a third-order parameter tensor, $\ell \in [L]$, and $L$ denotes the depth of the KAN. Therefore, the forward propagation of the KAN model can be formulated as
\begin{equation}
    \begin{split}
        \Psi\left(\bx;\left\{ \mathcal{A}_{\ell}\right\}_{\ell \in [L]}\right) & := \Phi_{L}\left( \Phi_{L-1}\ldots \left( \Phi_{1}\left(\bx; \mathcal{A}_{1} \right) ; \mathcal{A}_{L-1}\right) \mathcal{A}_{L} \right) \\
        & = \Phi_{L} \circ \Phi_{L-1} \circ \cdots \circ \Phi_{1} \left(\bm{x}; \left\{ \mathcal{A}_{\ell}\right\}_{\ell \in [L]} \right).
    \end{split}
\end{equation}
Suppose we first pre-train the KAN model on a given task, where the pre-trained model is denoted as $\Psi_{\text{pt}}\left(\bx;\left\{ \mathcal{A}_{\ell, \text{pt}}\right\}\right)$, and $\left\{ \mathcal{A}_{\ell, \text{pt}}\right\}$ represents the set of parameter tensors across layers. 
For a new task, the fine-tuned model is denoted as $\Psi_{\text{ft}}\left(\bx;\left\{ \mathcal{A}_{\ell,\text{ft}}\right\}\right)$, where $\left\{ \mathcal{A}_{\ell,\text{ft}}\right\}$ is the updated set of parameter tensors.
We define the target model, which represents the optimal function within the KAN function class, as $\Psi_{\text{tg}}\left(\bx;\left\{ \mathcal{A}_{\ell,\text{tg}}\right\}\right)$, where $\left\{ \mathcal{A}_{\ell,\text{tg}}\right\}$ denotes the corresponding set of target parameter tensors.
The parameter tensor $\mathcal{A}_{\ell,\text{ft}}$ of the fine-tuned model $\Psi_{\text{ft}}$ updated using LoTRA follows:
\begin{equation}
\label{eq_update_finetune}
    \mathcal{A}_{\ell,\text{ft}} = \mathcal{A}_{\ell,\text{pt}} + \mathcal{G}_{\ell} \times_{1} \bU_{\ell}^{(1)} \times_{2} \bU_{\ell}^{(2)} \times \bU_{\ell}^{(3)},
\end{equation}
where $\mathcal{G}_{\ell} \in \mathbb{R}^{r_{\ell,1} \times r_{\ell,2} \times r_{\ell,3}}$  denotes the core tensor, and $\bU_{\ell}^{(1)} \in \mathbb{R}^{n_{\ell} \times r_{\ell,1}}$, $\bU_{\ell}^{(2)} \in \mathbb{R}^{n_{\ell+1} \times r_{\ell,2}}$ and $\bU_{\ell}^{(3)} \in \mathbb{R}^{n_{d} \times r_{\ell,3}}$ represent transformation matrices. Here, we require that $r_{\ell,1} \leq n_{\ell}$, $r_{\ell,2} \leq n_{\ell+1}$, and $r_{\ell,3} \leq n_{d}$, where $(r_{\ell,1},r_{\ell,2},r_{\ell,3})$ is the core tensor size for $\ell$-th layer. 
Since the Tucker rank of the updates $(r_{\ell,1}, r_{\ell,2}, r_{\ell,3})$ is smaller than the original parameter size $(n_{\ell}, n_{\ell+1}, n_d)$, LoTRA effectively reduces the model complexity with integrated low-rank structures on tensor updates while preserving essential task-specific information in $\mathcal{A}_{\ell,\text{ft}}$.

We first pre-train the model $\Psi_{\text{pt}}$ on a given task, and obtain the corresponding parameter tensors $\left\{\mathcal{A}_{\ell,\text{pt}} \right\}_{\ell \in [L]}$.
When solving a new task that shares similarities with the pre-training task, we aim to transfer shared information to the new task. To achieve this, we fine-tune the model using LoTRA, resulting in the fine-tuned model $\Psi_{\text{ft}}$ with updated parameter tensors $\left\{\mathcal{A}_{\ell,\text{ft}} \right\}_{\ell \in [L]}$, following the update rule in \Cref{eq_update_finetune}.
In this adaptation, the shared information is retained in the pre-trained tensors $\left\{\mathcal{A}_{\ell,\text{pt}} \right\}_{\ell \in [L]}$ while the task-specific information is incorporated through the trainable parameters $\left\{\mathcal{G}_{\ell}, \bU_{\ell}^{(1)}, \bU_{\ell}^{(2)},\bU_{\ell}^{(3)} \right\}_{\ell \in [L]}$, which are the only parameters updated during fine-tuning in $\Psi_{\text{ft}}$.
By using this adaptation approach, LoTRA eliminates the need to re-learn shared information from data, thereby significantly improving training efficiency compared to training a new model from scratch. 
Moreover, the task-specific information is effectively extracted, maintaining model expressiveness and performances on new tasks. 
Moreover, instead of fine-tuning all parameters on the new task, the explicitly introduced low-rank structure in LoTRA acts as a regularization, mitigating overfitting and filtering out noise arising from limited training data.

\section{Theoretical Analysis}
In this section, we study the theoretical expressiveness of LoTRA within the framework of low Tucker-rank approximations. To enable efficient fine-tuning using LoTRA with gradient descent, we propose a theoretically grounded learning rate selection strategy. 
Moreover, we prove that assigning identical learning rates to all trainable parameters is inefficient for training. These theoretical results not only enhance our understanding of LoTRA’s expressiveness and capacity but also provide a deeper insight into the training process.

\subsection{Expressiveness of LoTRA}

We define the error tensor between the target parameter tensor and the pre-trained parameter tensor as:
\begin{equation*}
    \mathcal{E}_{\ell} = \mathcal{A}_{\ell,\text{tg}} - \mathcal{A}_{\ell,\text{pt}}.
\end{equation*}
Let $\bm{E}_{\ell}^{(1)} \in \mathbb{R}^{n_{\ell} \times (n_{\ell + 1} n_{d})}$, $\bm{E}_{\ell}^{(2)} \in \mathbb{R}^{n_{\ell + 1} \times (n_{\ell} n_{d})}$, and $\bm{E}_{\ell}^{(3)} \in \mathbb{R}^{n_{d} \times (n_{\ell} n_{\ell + 1})}$ denote the corresponding mode-$i$ unfolding matrices of $\mathcal{E}_{\ell}$. 
The following lemma on the Tucker approximation of tensors plays a crucial role in characterizing the approximation capability of LoTRA, which integrates low Tucker-rank adaptation to parameter tensor updates.

\begin{lemma}
[Tucker Approximation~\cite{de2000multilinear}] 
\label{lemma_tucker}
For each $\ell \in [L]$,
there exist $\left(\mathcal{G}_{\ell}, \bm{U}_{\ell}^{(1)}, \bm{U}_{\ell}^{(2)}, \bm{U}_{\ell}^{(3)}\right)$, such that
\begin{equation*}
\begin{split}
        \left\| \mathcal{A}_{\ell,\text{ft}} - \mathcal{A}_{\ell,\text{tg}} \right\|_{F}^{2} & \leq  \sum_{r = r_{\ell,1} + 1}^{n_{\ell}} \sigma_{r} \left(\bm{E}_{\ell}^{(1)} \right)^2\\
        & + \sum_{r = r_{\ell,2} + 1}^{n_{\ell + 1}} \sigma_{r} \left(\bm{E}_{\ell}^{(2)} \right)^2
        + \sum_{r = r_{\ell,3} + 1}^{n_{d}} \sigma_{r} \left(\bm{E}_{\ell}^{(3)} \right)^2,
\end{split}
\end{equation*}
where $\sigma_{r}(\cdot)$ denotes the $r$-th largest singular value of the given matrix, and   $\mathcal{A}_{\ell,\text{ft}}$ follows \Cref{eq_update_finetune}. 
\end{lemma}

 \begin{assumption}
\label{assump1}
    Assume that the basis functions $b_{k}\left( \cdot\right)$ are uniformly bounded and smooth, with Lipschiz constant $L >0$ and the bound $B > 0$. Moreover, the parameter tensors of both the target and pre-trained models are bounded by $M>0$, i.e., $\left\| \mathcal{A}_{\ell,\text{tg}}\right\|_{F} \leq M$ and $\left\| \mathcal{A}_{\ell,\text{pt}}\right\|_{F} \leq M$, for $\ell \in [L]$. 
\end{assumption}

The above assumption is mild and practically realizable for KAN models. The conditions of uniform boundedness and smoothness hold in common settings. For example, in KANs with B-spline basis functions, the input is restricted to a bounded domain, ensuring that the conditions are naturally satisfied~\cite{liu2025kan}. Similarly, for KANs using polynomial basis functions, such as Chebyshev and Legendre polynomials, the domain is usually normalized to \([-1,1]\) through an additional activation, such as the hyperbolic tangent function~\cite{ss2024chebyshev}. Therefore, the uniform boundedness and Lipschitz continuity conditions hold in practice. 
Based on \Cref{lemma_tucker} and \Cref{assump1}, we establish the approximation capability of LoTRA, particularly its ability to approximate the target function \(\Psi_{\text{tg}}\) using the fine-tuned model \(\Psi_{\text{ft}}\).

\begin{theorem}
\label{theorem_approximation}
    Under \Cref{assump1}, there exists a fine-tuned model $\Psi_{\text{ft}}$ with LoTRA model that satisfies 
    \begin{equation*}
        \begin{split}
            &  \quad \left\|\Psi_{\text{ft}}\left(\bm{x}\right) -  \Psi_{\text{tg}}\left(\bm{x}\right)\right\|_2 \\
            & \leq \sum_{\ell = 1}^{L} C_{\ell}\cdot \bigg( \sum_{r = r_{\ell,1} + 1}^{n_{\ell}} \sigma_{r} \left(\bm{E}_{\ell}^{(1)} \right)^2 + \sum_{r = r_{\ell,2} + 1}^{n_{\ell + 1}} \sigma_{r} \left(\bm{E}_{\ell}^{(2)} \right)^2 \\
            & \quad  + \sum_{r = r_{\ell,3} + 1}^{n_{d}} \sigma_{r} \left(\bm{E}_{\ell}^{(3)} \right)^2\bigg)^{1/2},
        \end{split}
    \end{equation*}
    for any $\bx \in \mathbb{R}^{n}$, where the constant $C_{\ell}:= \left(M L \sqrt{n_d} \right)^{L - \ell} \cdot B \sqrt{n_{\ell} n_{d}}$ depends only on the constants in \Cref{assump1} and the model size. 
\end{theorem}

The proof can be found in \Cref{appendix_proof1}.
\Cref{theorem_approximation} indicates that when some of the singular values of the error tensor $\mathcal{E}_{\ell}$ between the pre-trained and target models are negligible, the fine-tuned model with LoTRA is capable of accurately approximating the target model. The quality of this approximation depends on the magnitude of the discarded singular values.

\subsection{Efficient Training of LoTRA}

Besides the approximation capability of the fine-tuned model with LoTRA, optimization efficiency with gradient descent is also a key focus and a critical consideration for real-world applications. 
For analysis convenience, we consider fine-tuning the model on a toy case, where we are given one training data $(\bm{x},\bm{y})$ of a new task and the loss function is formulated as $\mathcal{L} = \frac{1}{2} \left\| \Psi_{\text{ft}}(\bm{x}) - \bm{y}\right\|_2^2$ with $\bx \in \mathbb{R}^{n}$ and $\bm{y} \in \mathbb{R}^{m}$. We first consider a one-layer model and the adaptation with core tensor size $(r_1, r_2, r_3)$. Suppose the pre-trained model is $\Psi_{\text{pt}}(\bx;\mathcal{A}_{\text{pt}})$ and the fine-tuned model $\Psi_{\text{ft}}(\bx;\mathcal{A}_{\text{ft}})$  satisfies $\mathcal{A}_{\text{ft}} = \mathcal{A}_{\text{pt}} + \mathcal{G} \times_{1} \bU^{(1)} \times_{2} \bU^{(1)} \times_{3} \bU^{(3)}$, where $\mathcal{A}_{\text{pt}} \in \mathbb{R}^{n \times m \times n_{d}}$, $\mathcal{G} \in \mathbb{R}^{r_1 \times r_2 \times r_3}$, $\bU^{(1)} \in \mathbb{R}^{n \times r_1}$, $\bU^{(2)} \in \mathbb{R}^{m \times r_2}$, and $\bU^{(3)} \in \mathbb{R}^{n_{d} \times r_1}$. Here, we assume that $(r_1,r_2,r_3)$ is fixed and much smaller than $(n,m,n_d)$. The pre-trained parameter tensor $\mathcal{A}_{\text{pt}}$ remains fixed and $\left\{\mathcal{G},\bU^{(1)},\bU^{(2)},\bU^{(3)}\right\}$ are trainable. 
We denote the $(p,q,k)$-th element of the tensor $\mathcal{G}$ as $g^{p,q,k}$ and the $j$-th column of $\bU^{(i)}=[\bu^{(i),1},\ldots,\bu^{(i),r_{i}}]$ is denoted as $\bu^{(i),j}$. All trainable parameters are optimized by the vanilla gradient descent, following the updates:
\begin{equation}
        g_{t}^{p,q,k} = g_{t-1}^{p,q,k} - \eta_{0} \cdot  \frac{\partial \mathcal{L}_{t-1}}{\partial g^{p,q,k}},\quad \bm{u}^{(i),j}_{t} =  \bm{u}^{(i),j}_{t-1} - \eta_{i} \cdot \frac{\partial \mathcal{L}_{t-1}}{\partial \bm{u}^{(i),j}},
\end{equation}
for $p \in [n]$, $q \in [m]$, $k \in [n_d]$, $j \in [r_{i}]$, and $i \in \{1,2,3\}$. Here, the subscript $t$ denotes the parameters after $t$ steps of gradient descent, and $(\eta_{0},\eta_{1},\eta_{2},\eta_{3})$ are learning rates for $\{\mathcal{G},\bU^{(1)},\bU^{(2)},\bU^{(3)}\}$ respectively. Similarly, we denote the fine-tuned model $\Psi_{\text{ft}}$ at step $t$ as $\Psi_{t,\text{ft}}$ with parameters  $\{\mathcal{G}_{t},\bU^{(1)}_{t},\bU^{(2)}_{t},\bU^{(3)}_{t}\}$, and the corresponding loss function is denoted as $\mathcal{L}_{t}$.

Denote $\bX \in \mathbb{R}^{n \times n_d}$, where the $k$-th column of $\bX$ is obtained by applying the basis function $b_{k}$ elementwise to the input column vector $\bx$. Then, the change in function values after one step of gradient descent satisfies
\begin{equation*}
    \begin{split}
        & \quad \Delta \Psi_{t,\text{ft}} := \Psi_{t,\text{ft}} - \Psi_{t-1,\text{ft}} \\
        & \approx \sum_{p=1}^{r_1} \sum_{q=1}^{r_2} \sum_{k=1}^{r_3} \delta_{t,0}^{p,q,k} + \sum_{p=1}^{r_1} \delta_{t,1}^{p} + \sum_{q=1}^{r_2} \delta_{t,2}^{q}  + \sum_{k=1}^{r_3} \delta_{t,3}^{k},
    \end{split}
\end{equation*}
where 
\begin{equation*}
    \begin{split}
        & \delta_{t,0}^{p,q,k} = - \eta_0 \cdot \left(\bm{v}_{t-1}^{\top} \bm{u}_{t-1}^{(2), q} \right) \cdot  \left( \bm{u}_{t-1}^{(1), p \top} \bm{X} \bm{u}_{t-1}^{(3),k}\right)^2 \cdot \bm{u}_{t-1}^{(2),q},\\
        & \delta_{t,1}^{p} = - \eta_{1} \cdot  \sum_{q=1}^{r_2} \sum_{k=1}^{r_3} g^{p,q,k}_{t-1} \left( \sum_{q^{\prime}=1}^{r_2} \sum_{k^{\prime}=1}^{r_3} g_{t-1}^{p,q^{\prime},k^{\prime}} \left(\bm{v}_{t-1}^{\top} \bm{u}_{t-1}^{(2), q^{\prime}} \right) \right. \\
        & \quad \quad \quad \left. \cdot \left(\bm{u}_{t-1}^{(3), k^{\prime} \top} \bm{X}^{\top} \bm{X} \bm{u}_{t-1}^{(3), k} \right) \right) \cdot \bm{u}_{t-1}^{(2),q},\\
        & \delta_{t,2}^{q} = - \eta_2 \cdot \sum_{p=1}^{r_1} \sum_{k=1}^{r_3} g^{p,q,k}_{t-1} \left( \sum_{p^{\prime} = 1}^{r_1} \sum_{k^{\prime}=1}^{r_3} g^{p^{\prime},q,k^{\prime}}_{t-1}\left( \bm{u}_{t-1}^{(1),p^{\prime} \top} \bm{X} \bm{u}_{t-1}^{(3), k^{\prime}}\right)\right) \\
        & \quad \quad \quad \cdot \left( \bm{u}_{t-1}^{(1),p \top} \bm{X} \bm{u}_{t-1}^{(3), k}\right) \cdot \bm{v}_{t-1},\\
        & \delta_{t,3}^{k} = - \eta_{3} \cdot \sum_{p=1}^{r_1} \sum_{q=1}^{r_2} g^{p,q,k}_{t-1} \left( \sum_{p^{\prime}=1}^{r_1}\sum_{q^{\prime}=1}^{r_2}  g_{t-1}^{p^{\prime},q^{\prime},k} \left(\bm{v}_{t-1}^{\top} \bm{u}_{t-1}^{(2), q^{\prime}} \right) \right. \\
        & \quad \quad \quad \left. \cdot \left(\bm{u}_{t-1}^{(1), p \top} \bm{X} \bm{X}^{\top} \bm{u}_{t-1}^{(1), p^{\prime}} \right) \right) \cdot \bm{u}_{t-1}^{(2),q},
    \end{split}
\end{equation*}
where $\bm{v}_{t} := \frac{\partial \mathcal{L}_{t}}{\partial \Psi_{\text{ft}}} = \Psi_{t,\text{ft}}(\bx) - \bm{y}$ denotes the partial derivative of the loss with respect to the output of the fine-tuned model. 
The detailed derivation of the linearization of $\Delta \Psi_{t,\text{ft}}$ can be found in \Cref{appendix_derivation1}
Here, $\{\delta_{t,0}^{p,q,k}, \delta_{t,1}^{p},\delta_{t,2}^{q},\delta_{t,3}^{k}\}$ represent the first-order linearization terms of $\Delta \Psi_{t,\text{ft}}$, and the higher-order terms with respect to learning rates are neglected. The training efficiency of LoTRA is evaluated based on the magnitude of the first-order improvement in the function value, as defined in \Cref{def_efficiency}.

\begin{definition}
\label{def_efficiency}
    The training of LoTRA is considered as efficient if$ \{ \|\delta_{t,0}^{p,q,k} \|_2  \}$, $ \{ \|\delta_{t,1}^{p} \|_2  \}$, $ \{ \|\delta_{t,2}^{q} \|_2  \}$, and $ \{ \|\delta_{t,3}^{k} \|_2  \}$ are $ \Theta (1 )$ with respect to the model size $(n,m,n_d)$, for $p \in [r_1]$, $q \in [r_2]$, $k \in [r_3]$, and $t \geq 2$.
\end{definition}

The definition of training efficiency assumes that the first-order improvement in function values captures the training progress, as learning rates are typically set to be small. Efficient training requires that the one-step improvement remains independent of the model size. 
To illustrate this, consider a scenario where the one-step improvement follows the order $\Theta\left( n^{\alpha_1} m^{\alpha_2} n_d^{\alpha_3}\right)$. For stable training, it must hold that $\alpha_1, \alpha_2, \alpha_3 \leq 0$, otherwise, the model exhibits value explosion, leading to instability and training failure. However, if $\alpha_1, \alpha_2, \alpha_3 <0$, the one-step improvement scales negatively with model size, resulting in excessively slow training as the model scales up. This is undesirable in practice. Therefore, to achieve both stability and efficiency, we prefer the condition $\alpha_1 = \alpha_2 = \alpha_3 = 0$, as stated in \Cref{def_efficiency}.

Before delving into the details of the analysis, it is crucial to consider the initialization setup of parameters, because the initialization together with the learning rates determines the scale of change of function values during training. We initialize all trainable parameters as follows:
\begin{align}
     & g_0^{p,q,k} = 0, \quad \bm{u}_{0}^{(1),q} \sim \mathcal{N}\left(\bm{0},\frac{1}{n} \bm{I}\right),\quad \bm{u}_{0}^{(3),k} \sim \mathcal{N}\left(\bm{0},\frac{1}{n_{d}} \bm{I}\right), \nonumber\\
    & \left\|\bm{u}_{0}^{(2),q}\right\|_2 = \Theta\left(1 \right),~\text{and}~ \bm{u}_{0}^{(2),q \top} \bm{v}_{0} =  \Theta\left(1 \right),\label{eq_initialization}
\end{align}
for $p \in [r_1]$, $q \in [r_2]$ and $k \in [r_3]$, where $\bm{v}_{0} := \frac{\partial \mathcal{L}_{0}}{\partial \Psi_{\text{ft}}} = \Psi_{0,\text{ft}}(\bx) - \bm{y} = \Psi_{\text{pt}}(\bx) - \bm{y} $ denotes the partial derivative of the loss with respect to the output of the fine-tuned model at initialization. 
The first three initialization schemes are standard, except for the initialization of $\bu_{0}^{(2),q}$. 
A natural choice that satisfies the given condition is to explicitly compute the vector  $\bm{v}_{0}$ based on the pre-trained model and initialize all $\bu_{0}^{(2),q}$ as $\bm{v}_{0}/ \left\| \bm{v}_{0}\right\|_2$.

The following theorem presents the main result of the efficient training analysis for LoTRA. The proof extends the work of Hayou et al.~\cite{hayou2024lora}, 
who analyzed the efficient training of LoRA involving two parameter components. In contrast, the proposed LoTRA framework is significantly more complex, consisting of four parameter components. Therefore, extending the proof and analysis is nontrivial and requires substantial modifications.

\begin{theorem}
\label{theorem_efficient}
    Efficient training of a one-layer KAN with LoTRA is not achievable, if all learning rates are set to the same order of magnitude with respect to the model size
    $(n,m,n_d)$, under the initialization scheme in \Cref{eq_initialization}. 
    However, efficient training can be achieved, if the learning rates are set as follows: $\eta_0 = \Theta\left(1\right)$, $\eta_1 = \Theta\left(n^{-1}\right)$, $\eta_2 = \Theta\left(1\right)$, and $\eta_3 = \Theta\left(n_d^{-1}\right)$.
\end{theorem}

The proof can be found in \Cref{appendix_proof2}.
\Cref{theorem_efficient} provides theoretical insights into learning rate selection for LoTRA. As the input dimension increases, the learning rate $\eta_{1}$ for the mode-1 transformation matrix should be decreased accordingly. Similarly, the learning rate $\eta_3$ for the mode-3 transformation matrix should be adjusted based on the number of basis functions $n_d$. 
In contrast, the learning rates $\eta_0$ for the core tensor and $\eta_{2}$ for the mode-2 transformation matrix remain independent of the model size. This theorem highlights an important principle for learning rate selection, providing valuable practical guidance for learning rate tuning.
In practice, the hidden dimension (or input dimension)  $n$ is usually larger than the number of basis functions $n_d$, i.e., $n > n_d$. Based on our theoretical results, an appropriate learning rate setup might follow $\eta_0 \approx \eta_2 \geq \eta_3 > \eta_{1}$.
These theoretical results significantly streamline the learning rate selection process, reducing the learning rate search space and facilitating a more efficient tuning.

The extension of \Cref{theorem_efficient} to multi-layer KANs is feasible. The definition of efficient training in \Cref{def_efficiency} can be adapted for multi-layer KANs by requiring that all first-order linearization terms of each layer remain $\Theta(1)$ with respect to the model size. The proof follows a similar structure to that of \Cref{theorem_efficient}, with minor modifications on $\bm{v}_{t}^{\ell} := \frac{\partial \mathcal{L}}{\bz_{\ell,\text{ft}}}$, which represents the partial derivative of the loss function with respect to the output of $\ell$-th layer (denoted as $\bz_{\ell,\text{ft}}$) of the fine-tuned model $\Psi_{t,\text{ft}}$.
The initialization of all trainable parameters is given by
\begin{equation*}
\begin{split}
    & g_0^{\ell,p,q,k} = 0,~\bm{u}_{0}^{(1),\ell,q} \sim \mathcal{N}\left(\bm{0},\frac{1}{n_{\ell}} \bm{I}\right),~\bm{u}_{0}^{(3),\ell,k} \sim \mathcal{N}\left(\bm{0},\frac{1}{n_{d}} \bm{I}\right),\\
    & \left\|\bm{u}_{0}^{(2),\ell,q}\right\|_2 = \Theta\left(1 \right),~\text{and}~ \bm{v}_{0}^{\ell \top}\bm{u}_{0}^{(2),\ell,q} =  \Theta\left(1 \right),
\end{split}
\end{equation*}
for $p \in [r_1]$, $q \in [r_2]$, $k \in [r_3]$, and $\ell \in [L]$. 
Here, the initialization for $\bm{u}_{0}^{(2),\ell,q}$ can be achieved by explicitly calculating $\bm{v}_{0}^{\ell}$ for each layer based on the fine-tuned model, and setting all $\bm{u}_{0}^{(2),\ell,q}$ to be $\bm{v}_{0}^{\ell}/\left\|\bm{v}_{0}^{\ell} \right\|_2$.
\Cref{corollary_efficient} establishes that the theoretically optimal learning rates for each layer depend on both the input dimension and the number of basis functions.

\begin{corollary}
\label{corollary_efficient}
    Efficient training of multi-layer KANs with LoTRA is not achievable, if all learning rates are set to the same order of magnitude as the model size $\{n_{\ell}\}_{\ell \in [L]}$ and $n_d$. However, for the $\ell$-th layer of the fine-tuned model, efficient training holds if the learning rates are chosen as follows: $\eta_0 = \Theta\left(1\right)$, $\eta_1 = \Theta\left(n_{\ell}^{-1}\right)$, $\eta_2 = \Theta\left(1\right)$, and $\eta_3 = \Theta\left(n_d^{-1}\right)$.
\end{corollary}

\section{Applications}
In this section, we discuss the potential applications of LoTRA in fine-tuning and training KANs.

\subsection{Physics-Informed KANs}
\label{sec_pikans}
KANs have demonstrated superior performance over MLPs in some science-related tasks, making them promising models for physics-informed machine learning. We investigate the potential of LoTRA in training KANs for solving a class of PDEs, aiming to reduce computational cost and storage requirements with enhanced efficiency.  
Consider the problem of solving a class of PDEs with different physical parameters $\blam \in \mathcal{S}$:
\begin{equation*}
    \begin{split}
        & \mathcal{D}[\bu_{\blam};\blam] = f(\bx;\blam),\quad\bx \in \Omega, \\
        & \mathcal{B}[\bu_{\bm{\lambda}};\bm{\lambda}] = g(\bx;\blam),\quad\bx \in \partial \Omega, 
    \end{split}
\end{equation*}
where $\mathcal{D}$ and $\mathcal{B}$ are differential operators defined in the interior domain $\Omega$ and on its boundary $\partial \Omega$, respectively. 
The solution corresponding to physical constants $\blam$ is denoted as $\bu_{\blam}$.

In physics-informed machine learning, neural networks act as surrogates for PDE solutions. Given a KAN model $ \Psi\left(\bx;\left\{ \mathcal{A}_{\ell}\right\}_{\ell \in [L]}\right) $, the training loss is formulated as
\begin{align}
       & \mathcal{L}\left(\left\{ \mathcal{A}_{\ell}\right\}_{\ell \in [L]}\right)   = \frac{\mu}{N} \sum_{i=1}^{N} \left\| \mathcal{D}\left[\Psi\left(\bx_{i};\left\{ \mathcal{A}_{\ell}\right\}_{\ell \in [L]}\right)\right] - \by_{i} \right\|_2^2 \nonumber \\*
        & \qquad\qquad + \frac{\mu_{b}}{N_{b}} \sum_{j=1}^{N_b} \left\| \mathcal{B}\left[\Psi\left(\hat{\bx}_{j};\left\{ \mathcal{A}_{\ell}\right\}_{\ell \in [L]}\right)\right] - \bb_{j} \right\|_2^2,
\end{align}
given some observations $\left\{ (\bx_{i},\by_{i})\right\}_{i \in [N]}$ with $\by_{i} = f(\bx_{i};\blam)$ in the interior and $\left\{ (\hat{\bx}_{j},\bb_{j})\right\}_{j \in [N_{b}]}$ with $\bb_{j} = g(\hat{\bx}_{j};\blam)$ on the boundary. Here, $\mu >0$ and $\mu_{1}>0$ are hyperparameters that balance the PDE residual in the interior domain and the boundary condition residual, respectively.

Our objective is to train KANs to efficiently approximate solutions $\{\bu_{\blam}\}_{\blam \in \mathcal{S}}$ for the entire class of PDEs. 
Due to structural similarities in the PDE operators, we assume that the solutions exhibit shared underlying patterns. Instead of repeatedly learning this shared information from each individual PDE, we adopt LoTRA to retain the shared information across all PDEs and adapt to task-specific variations using a low tensor-rank adaptation framework. This enables efficient fine-tuning of different PDEs, reducing redundancy in training and significantly improving storage efficiency.

Given a PDE with physical parameters $\blam_{0}$, we first pre-train the full KAN model on this PDE, where the obtained model $\Psi_{\text{pt}}$ is an approximation to the corresponding solution $\bu_{\blam_{0}}$. The shared information across the class of PDEs is inherently captured in $\Psi_{\text{pt}}$. 
For a new PDE with different physical parameters $\hat{\blam} \in \mathcal{S}$, we fine-tune the KANs model by LoTRA and obtain $\Psi_{\text{ft}}$, which approximates the solution $\bu_{\hat{\blam}}$. 
The fine-tuning process is efficient because the shared information has already been included, eliminating the need for redundant learning, and thus resulting in faster convergence.

Moreover, when storing KAN models for an entire class of PDEs, the storage requirements are significantly reduced if the core tensor $\mathcal{G}$ is much smaller than the original parameter tensor. In this case, rather than storing the full model for each PDE, it suffices to retain the parameter tensors of $\Psi_{\text{pt}}$ and the low tensor-rank components $\big\{\big(\mathcal{G}_{\ell}, \bm{U}_{\ell}^{(1)}, \bm{U}_{\ell}^{(2)}, \bm{U}_{\ell}^{(3)}\big)\big\}_{\ell \in [L]}$ for each PDE. 
For notational simplicity, we use ``cr'' to denote the compression ratio of LoTRA. 
A compression ratio of $\text{cr} = 1/4$ implies that each mode of the core tensor $\mathcal{G}$ is compressed by $1/4$, reducing the dimensionality of each mode and resulting in an overall parameter ratio of $(1/4)^{3}$ compared to the original parameter tensor. Mathematically, this is expressed as $(r_{\ell,1},r_{\ell,2},r_{\ell,3}) = \lceil \text{cr} \cdot (n_{\ell},n_{\ell+1},n_d) \rceil$, where the ceiling operator $\lceil \cdot \rceil $ is applied elementwise to ensure that each component is rounded up to the nearest integer.
This approach significantly reduces parameter size while maintaining the expressiveness and adaptability of the model.

\subsection{Slim KANs}
Another potential application of LoTRA is slimming the model, enabling a slimmer KAN architecture with low tensor-rank structures on parameter tensors. Unlike transfer learning, this approach directly imposes a low tensor-rank structure on the parameter tensors of KANs from the outset. 
Suppose no pre-trained model is available, and we initialize all pre-trained parameter tensors $\mathcal{A}_{\ell,\text{pt}}$ as zero tensors.
This setup is equivalent to assuming that the parameter tensors themselves exhibit low tensor-rank properties. Similar assumptions have been validated in other domains, such as large language models~\cite{aghajanyan2020intrinsic}. Additionally, in \Cref{sec_motivation}, we empirically demonstrate that KAN parameter tensors naturally exhibit a low tensor-rank structure in the function representation task.   
In this case, the only trainable parameters in the slim KAN model are  $\big\{\big(\mathcal{G}_{\ell}, \bm{U}_{\ell}^{(1)}, \bm{U}_{\ell}^{(2)}, \bm{U}_{\ell}^{(3)}\big)\big\}_{\ell \in [L]}$.
If the core tensor is significantly smaller than the full parameter tensor, it results in a much smaller model with certain expressiveness, compared to the full KAN with the same width and depth.  
Besides the benefits of reduced parameter size, the integrated low-rank structure of parameters acts as a form of regularization, preventing overfitting and improving generalization. 
In our experiments, we applied the slim KAN model to function representation and image classification tasks. The results show that it achieves comparable performance to vanilla KANs, demonstrating the potential of slim KANs constructed by LoTRA.

\section{Experiments}
In this section, we conduct comprehensive experiments on transfer learning of KANs using LoTRA in solving a class of PDEs. 
Additionally, we validate the learning rate selection strategy derived in \Cref{theorem_efficient}. 
Slim KANs are also evaluated and compared with MLPs on function representation and image classification tasks. 
To explore the impact of different basis functions, we consider Chebyshev polynomials, Legendre polynomials, Taylor polynomials, and Fourier series as basis functions, and denote the corresponding KAN models as ChebyKAN, LegendreKAN, TaylorKAN, and FourierKAN, respectively.  
For all pre-training and fine-tuning models, we adopt the Adam optimizer with its default hyperparameters.

\subsection{Transfer Learning of KANs}
In this experiment, we apply KANs with LoTRA to solve a class of PDEs, as detailed in \Cref{sec_pikans}. We consider three types of second-order PDEs: elliptic, parabolic, and hyperbolic equations. 
For a given PDE in the class, we first pre-train a KAN model to obtain $\Psi_{\text{pt}}$. For new PDE tasks within the same class, we fine-tune the pre-trained model using LoTRA, resulting in the fine-tuned model $\Psi_{\text{ft}}$.  
To evaluate the quality of the obtained models, we compute the relative error (rel) between the predicted solution $\Psi(\bx)$ and the exact solution $\bu(\bx)$. Given a test dataset $\left\{(\bx_{i},\bu(\bx_{i}))\right\}_{i \in N_{t}}$, the relative error is defined as 
$
    \text{rel} = \frac{\sum_{i=1}^{N_{t}} \left\|\Psi(\bx_{i}) - \bu(\bx_{i}) \right\|_2^2 }{\sum_{i=1}^{N_{t}} \left\| \bu(\bx_{i}) \right\|_2^2}.
$
In the figures, the method labeled ``full'' refers to the vanilla transfer learning approach, where all parameters are fully updated. The method labeled ``zero'' represents training from scratch without leveraging any information from a pre-trained model.

As discussed in \Cref{sec_pikans}, we denote the compression ratio of LoTRA as ``cr."  
A compression ratio of $\text{cr} = 1/4$ implies that each mode of the core tensor $\mathcal{G}$ is reduced to $1/4$ of its original size, leading to an overall parameter reduction of $(1/4)^{3}$ compared to the original parameter tensor. The mathematical formulation follows $(r_{\ell,1},r_{\ell,2},r_{\ell,3}) = \lceil \text{cr} \cdot (n_{\ell},n_{\ell+1},n_d) \rceil$, where the ceiling operator $\lceil \cdot \rceil $ is applied elementwise to ensure that each component is rounded up to the nearest integer.
In our experiments, we utilize LoTRA with three compression ratios, namely $\text{cr} = 1/2$, $1/4$, and $1/8$. 
We adopt three-layer KANs with a fixed hidden dimension $n_{\ell}=64$ and the number of basis functions set to $n_d = 8$. We consider PDEs with two-dimensional spatial domains. Since the input dimension and the output dimension studied here are significantly smaller than the hidden dimension, we apply LoTRA only to the hidden layer, while fine-tuning the input and the output layers normally without compression. 

\subsubsection{Elliptic Equations}
\label{sec_transfer_elliptic}
We consider a class of elliptic equations with varying parameters $\lambda \in \mathbb{R}$:
\begin{align}
        - \nabla \cdot \left(a(\bm{x}) \cdot \nabla u(\bm{x};\lambda)\right) + \left\| \nabla u(\bm{x};\lambda)\right\|_2^2   = f(\bm{x};\lambda), \quad \bm{x} &\in \Omega, \nonumber\\*
        u(\bm{x};\lambda) = g(\bm{x};\lambda), \quad \bm{x} &\in \partial \Omega,
       \end{align} 
where the domain is defined as $\Omega = \{\bx \in \mathbb{R}^{2}: \left\|\bx \right\|_2 \leq 1\}$, and the coefficient function is given by $a(\bx) = 1 + \frac{1}{2}\left\| \bx\right\|_2^2$. The exact solution $\bu(\bx;\lambda)$ with the parameter $\lambda$ is defined as $u(\bx;\lambda) = \sin \big(\frac{\pi}{2}\big( 1 - \left\|\bm{x} \right\|_2 \big)^{2.5} \big) + \lambda \cdot \sin \big(\frac{\pi}{2}\big( 1 - \left\|\bm{x} \right\|_2 \big) \big)$. 
We first pre-train a KAN model on the PDE with $\lambda =0$, and then fine-tune it on PDEs with $\lambda=0.1$ and $\lambda = 1$.

We first validate the developed theorem on the learning rate selection by designing four different learning rate strategies. In the first strategy, we set the learning rates based on \Cref{theorem_efficient} considering the hidden dimension and the number of basis functions. Specifically, we choose $(\eta_{0},\eta_{1},\eta_{2},\eta_{3}) = (\text{1e-2, 2e-4, 1e-2, 1e-3})$, which is consistent with the theorem if the constant in $\Theta(\cdot)$ is assumed to be 1e-2. For the second and third strategies, we set learning rates for all components of LoTRA to be 3e-3 and 1e-3, respectively. In the fourth strategy, we deliberately violate the theoretical conditions by making $\eta_{1}$ the largest learning rate. Specifically, we set $(\eta_{0},\eta_{1},\eta_{2},\eta_{3}) = (\text{2e-4, 1e-2, 1e-3, 1e-2})$. 
The fine-tuning trajectories of KANs with LoTRA under these learning rates strategies are illustrated in Figure \ref{fig_cheby_legendre_lr_elliptic} (as well as in Figures S1 and S3 in the supplementary material). 
In these figures, ``LR-1" to ``LR-4" correspond to the four respective learning rate strategies.
Across different transfer learning tasks, including varying parameter values $\lambda$, different compression ratios, and various KAN variants, we observe that the first strategy, designed from our theoretical results, achieves the lowest training loss in most cases. This empirical observation validates \Cref{theorem_efficient}. 
In some cases, the first strategy exhibits slightly lower performance. This is primarily due to the elliptic PDE examples being relatively easy for KANs to learn, leading to an overly simple fine-tuning process where differences between strategies are not as pronounced.

We compare the fine-tuned KANs using the introduced LoTRA with several baseline methods: the vanilla transfer learning approach, which updates all parameters based on the pre-trained model; KANs trained from scratch without transfer learning; and MLPs using the vanilla transfer learning method. The visualized results are shown in Figure \ref{fig_cheby_legendre_elliptic} (as well as in Figures S2 and S4 in the supplementary material).
The results demonstrate that KAN models with LoTRA are highly competitive and perform comparably with the vanilla transfer learning method updating all parameters, with significantly reduced parameter size and maintained performance.
We further observe that compression ratios of $\text{cr}=1/2$ and $\text{cr}=1/4$ exhibit comparable performances, with the latter occasionally outperforming the former. This is likely due to the simplicity of the example, where the compression ratio remains too large for $\text{cr}=1/2$, making $\text{cr}=1/4$ a suitable compression ratio for this case.  
To further evaluate performance, we compute the relative error of the fine-tuned models, as shown in Tables \ref{tab_eps0.1_elliptic} and \ref{tab_eps1.0_elliptic}. The results indicate that fine-tuned models with LoTRA perform comparably to, and in some cases outperform, full fine-tuning models. 
Additionally, KANs consistently outperform MLPs, highlighting their potential in physics-informed machine learning tasks.

\begin{figure*}[!t]
\centering
    \begin{minipage}[t]{0.3\textwidth} 
    \centering
    \subfigure[$\epsilon=0.1$, $\text{cr}=1/8$]{%
    \includegraphics[width=\textwidth]{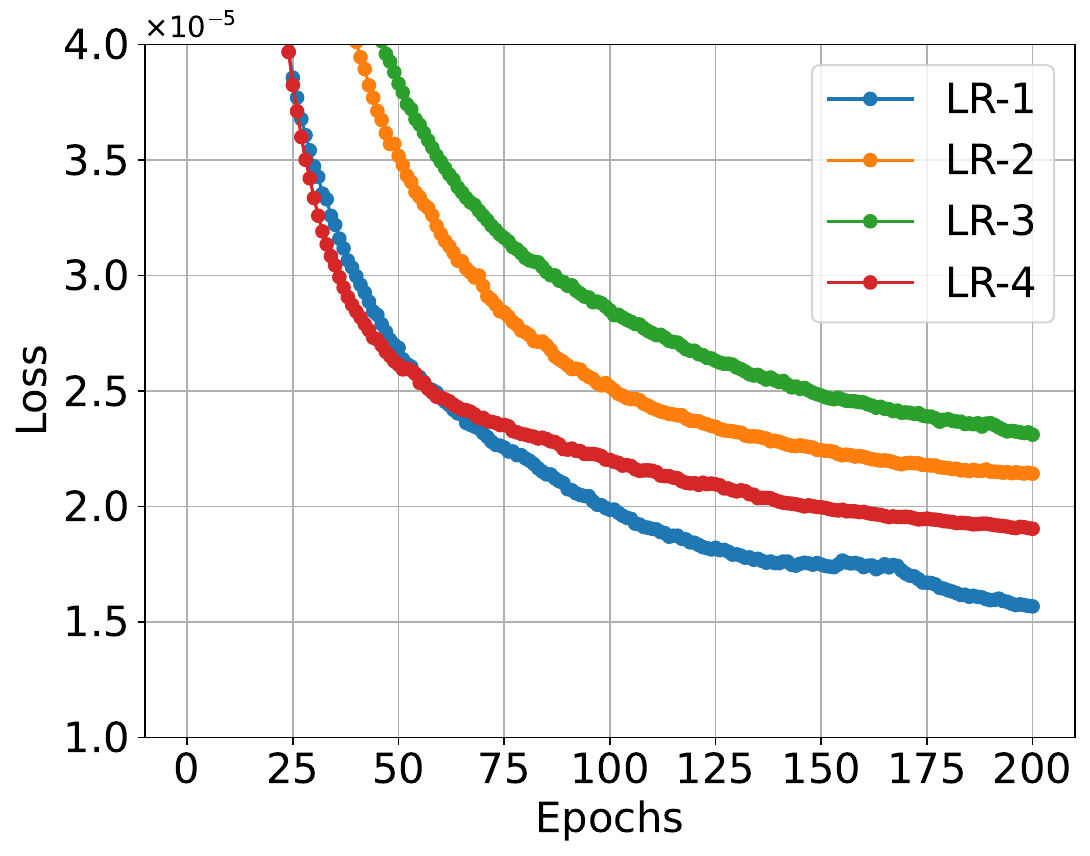} 
    }
    \caption{Fine-tuning trajectories of Chebyshev KANs using LoTRA under four strategies of learning rate selection (denoted as ``LR-1" to ``LR-4") for solving elliptic equations.}
    \label{fig_cheby_legendre_lr_elliptic}
    \end{minipage}
    \hfill
    \begin{minipage}[t]{0.66\textwidth}
    \centering
    \subfigure[$\epsilon=0.1$]{%
        \includegraphics[width=0.48\textwidth]{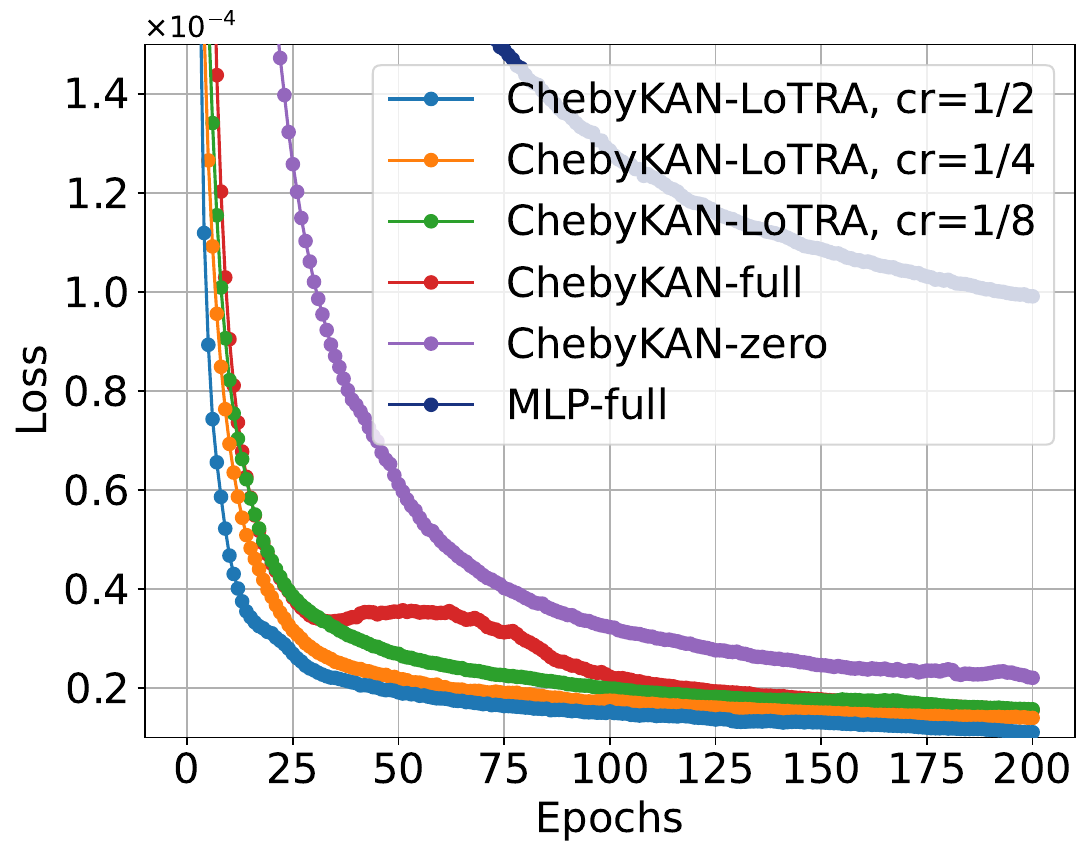} 
        }
    \subfigure[$\epsilon=1.0$]{%
        \includegraphics[width=0.48\textwidth]{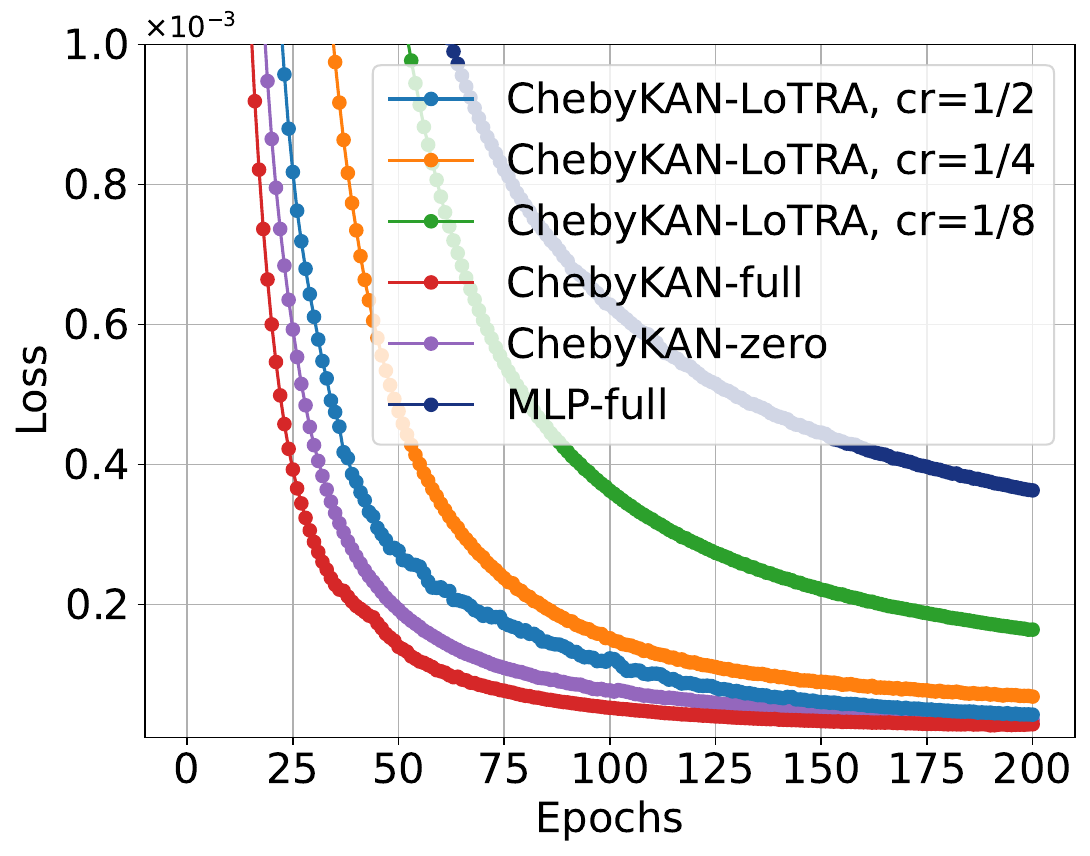} 
        }
    \caption{Fine-tuning trajectories of Chebyshev KANs using LoTRA, compared to fully updated KANs and MLPs, for solving elliptic equations.}
    \label{fig_cheby_legendre_elliptic}
    \end{minipage}
\end{figure*}

\begin{table}[ht]
\centering
\begin{tabular}{|ll|l|ll|l|}
\hline
\multicolumn{2}{|l|}{Method}                                      & \multirow{2}{*}{rel (\%)} & \multicolumn{2}{l|}{Method}                                     & \multirow{2}{*}{rel (\%)} \\ \cline{1-2} \cline{4-5}
\multicolumn{1}{|l|}{Type}                          & cr &                                    & \multicolumn{1}{l|}{Type}                         & cr &                                    \\ \hline
\multicolumn{1}{|l|}{\multirow{2}{*}{ChebyKAN}} & 1/2 &  0.05 & \multicolumn{1}{l|}{\multirow{2}{*}{TaylorKAN}} & 1/2 & 0.20  \\ \cline{2-3} \cline{5-6} 
\multicolumn{1}{|l|}{\multirow{2}{*}{-LoTRA}} & 1/4 & 0.06 & \multicolumn{1}{|l|}{\multirow{2}{*}{-LoTRA}} &  1/4  & 0.25  \\ \cline{2-3} \cline{5-6} 
\multicolumn{1}{|l|}{} &  1/8  & 0.08  & \multicolumn{1}{l|}{} & 1/8 & 0.23  \\ \cline{2-3} \cline{5-6}
\hline
\multicolumn{2}{|l|}{ChebyKAN-full} & 0.05  & \multicolumn{2}{|l|}{TaylorKAN-full} & 0.26 \\ \cline{2-3} \cline{5-6} 
\hline
\multicolumn{2}{|l|}{ChebyKAN-zero} & 0.08  & \multicolumn{2}{|l|}{TaylorKAN-zero} & 1.34 \\ \cline{2-3} \cline{5-6}
\hline
\multicolumn{1}{|l|}{\multirow{2}{*}{LegendreKAN}} & 1/2 &  0.08  & \multicolumn{1}{l|}{\multirow{2}{*}{FourierKAN}} & 1/2 & 0.14 \\ \cline{2-3} \cline{5-6} 
\multicolumn{1}{|l|}{\multirow{2}{*}{-LoTRA}} & 1/4 &  0.08 & \multicolumn{1}{l|}{\multirow{2}{*}{-LoTRA}} & 1/4 & 0.11  \\ \cline{2-3} \cline{5-6} 
\multicolumn{1}{|l|}{}  & 1/8 & 0.09 & \multicolumn{1}{l|}{}  &  1/8 &  0.13   \\ \cline{2-3} \cline{5-6} 
\hline
\multicolumn{2}{|l|}{LegendreKAN-full} & 0.08  & \multicolumn{2}{|l|}{FourierKAN-full} & 0.14 \\ \cline{2-3} \cline{5-6}
\hline
\multicolumn{2}{|l|}{LegendreKAN-zero} & 0.14  & \multicolumn{2}{|l|}{FourierKAN-zero} & 0.22 \\ \cline{2-3} \cline{5-6}
\hline
\multicolumn{2}{|l|}{MLP-full} & 0.30 & \multicolumn{2}{l|}{MLP-zero}                          &  0.80 \\ \hline
\end{tabular}
\caption{Relative error (rel)  of KANs with LoTRA, fully updated KANs and MLPs, and KANs and MLPs trained from scratch without transfer learning, for solving elliptic equations with parameter $\epsilon=0.1$.}
\label{tab_eps0.1_elliptic}
\end{table}

\begin{table}[htb!]
\centering
\begin{tabular}{|ll|l|ll|l|}
\hline
\multicolumn{2}{|l|}{Method}                                      & \multirow{2}{*}{rel (\%)} & \multicolumn{2}{l|}{Method}                                     & \multirow{2}{*}{rel (\%)} \\ \cline{1-2} \cline{4-5}
\multicolumn{1}{|l|}{Type}                          & cr &                                    & \multicolumn{1}{l|}{Type}                         & cr &                                    \\ \hline
\multicolumn{1}{|l|}{\multirow{2}{*}{ChebyKAN}} & 1/2 &  0.08 & \multicolumn{1}{l|}{\multirow{2}{*}{TaylorKAN}} & 1/2 & 0.19  \\ \cline{2-3} \cline{5-6} 
\multicolumn{1}{|l|}{\multirow{2}{*}{-LoTRA}} & 1/4 & 0.09 & \multicolumn{1}{|l|}{\multirow{2}{*}{-LoTRA}} &  1/4  & 0.22  \\ \cline{2-3} \cline{5-6} 
\multicolumn{1}{|l|}{} &  1/8  & 0.10  & \multicolumn{1}{l|}{} & 1/8 & 0.20 \\ \cline{2-3} \cline{5-6}
\hline
\multicolumn{2}{|l|}{ChebyKAN-full} & 0.04  & \multicolumn{2}{|l|}{TaylorKAN-full} & 0.19 \\ \cline{2-3} \cline{5-6} 
\hline
\multicolumn{2}{|l|}{ChebyKAN-zero} & 0.04  & \multicolumn{2}{|l|}{TaylorKAN-zero} & 0.55 \\ \cline{2-3} \cline{5-6}
\hline
\multicolumn{1}{|l|}{\multirow{2}{*}{LegendreKAN}} & 1/2 &  0.14  & \multicolumn{1}{l|}{\multirow{2}{*}{FourierKAN}} & 1/2 & 0.16 \\ \cline{2-3} \cline{5-6} 
\multicolumn{1}{|l|}{\multirow{2}{*}{-LoTRA}} & 1/4 &  0.20 & \multicolumn{1}{l|}{\multirow{2}{*}{-LoTRA}} & 1/4 & 0.17  \\ \cline{2-3} \cline{5-6} 
\multicolumn{1}{|l|}{}  & 1/8 & 0.33 & \multicolumn{1}{l|}{}  &  1/8 &  0.39   \\ \cline{2-3} \cline{5-6} 
\hline
\multicolumn{2}{|l|}{LegendreKAN-full} & 0.10  & \multicolumn{2}{|l|}{FourierKAN-full} & 0.10 \\ \cline{2-3} \cline{5-6}
\hline
\multicolumn{2}{|l|}{LegendreKAN-zero} & 0.15  & \multicolumn{2}{|l|}{FourierKAN-zero} & 0.18 \\ \cline{2-3} \cline{5-6}
\hline
\multicolumn{2}{|l|}{MLP-full} & 0.25 & \multicolumn{2}{l|}{MLP-zero}                          &  0.28 \\ \hline
\end{tabular}
\caption{Relative error (rel)  of KANs with LoTRA, fully updated KANs and MLPs, and KANs and MLPs trained from scratch without transfer learning, for solving elliptic equations with parameter $\epsilon=1.0$.}
\label{tab_eps1.0_elliptic}
\end{table}

\subsubsection{Allen-Cahn Equations}
\label{sec_transfer_allen_cahn}
We consider a class of Allen-Cahn equations, which are nonlinear parabolic PDEs, with varying parameters $\lambda \in \mathbb{R}$:
\begin{equation}
    \begin{split}
        \frac{\partial u(t,\bm{x};\lambda)}{\partial t} & - \Delta u(t,\bm{x};\lambda) -u(t,\bm{x};\lambda)^3 + u(t,\bm{x};\lambda)\\
        & = f(t,\bm{x};\lambda), \quad (t,\bm{x}) \in [0,1] \times \Omega,\\
        u(t,\bm{x};\lambda) & = g(t,\bm{x};\lambda), \quad (t,\bm{x}) \in [0,1] \times \partial \Omega,\\
        u(0,\bm{x};\lambda) & = h(\bm{x};\lambda), \quad \bm{x} \in \Omega,
    \end{split}
\end{equation}
where the temporal and the spatial domains are defined as $[0,1]$ and $\Omega = \{\bx \in \mathbb{R}^{2}: \left\|\bx \right\|_2 \leq 1\}$, respectively. The exact solution $\bu(\bx;\lambda)$ with the parameter $\lambda$ is defined as $u(\bx;\lambda) = e^{-t} \sin \left(\frac{\pi}{2}\left( 1 - \left\|\bm{x} \right\|_2\right)^{2.5} \right) + \lambda \cdot e^{-t} \sin \left(\frac{\pi}{2}\left( 1 - \left\|\bm{x} \right\|_2\right) \right)$. 
We first pre-train a KAN model on the PDE with $\lambda =0$, and then fine-tune it on PDEs with $\lambda=0.1$ and $\lambda = 1$.

We conduct similar experiments as in \Cref{sec_transfer_elliptic} for solving Allen-Cahn equations using KANs with LoTRA, adopting four different learning rate strategies to verify the derived \Cref{theorem_efficient}. 
The four strategies follow those in  \Cref{sec_transfer_elliptic}, where the first strategy is consistent with the theorem, while the other three deviate from the theoretically optimal learning rate selection.   
The fine-tuning trajectories of KANs with LoTRA under these learning rates strategies are illustrated in Figures \ref{fig_cheby_legendre_lr_allen_cahn} (as well as in Figures S5 and S7 in the supplementary material).
Across different transfer learning tasks, including varying parameter values $\lambda$, different compression ratios, and various KAN variants, we observe that the first strategy, derived from our theoretical results, consistently achieves the lowest training loss in all cases. This empirical observation validates \Cref{theorem_efficient}, demonstrating the effectiveness of the proposed learning rate selection strategy. 
The advantages of using the first strategy are particularly obvious in tasks with $\epsilon=1.0$, as these cases are relatively more challenging. Therefore, the benefits of the theoretically guided learning rate selection become more noticeable compared to the results obtained for elliptic equations in Figure \ref{fig_cheby_legendre_lr_elliptic}.

We compare the fine-tuned KANs using the introduced LoTRA with several baseline methods for solving Allen-Cahn equations. The baseline models are the same as those examined in \Cref{sec_transfer_elliptic}. The visualized results are presented in Figures \ref{fig_cheby_legendre_allen_cahn} (as well as in Figures S6 and S8 in the supplementary material).
The results demonstrate that KAN models with LoTRA outperform and sometimes comparably with the vanilla fine-tuning method with full parameter updates. Additionally, they significantly outperform models trained from scratch without transfer learning.  
We further observe that LoTRA with larger compression ratios (i.e., larger core tensors) achieve lower training loss, benefiting from greater expressiveness.
This observation differs slightly from the results for elliptic equations, mainly because Allen-Cahn equations are more challenging to solve. Therefore, when applying LoTRA, it is crucial to balance the trade-off between compression ratio and model expressiveness, depending on the specific application requirements.  
To further evaluate performance, we compute the relative error of the fine-tuned models, as shown in Tables SI and SII in the supplementary material. The results indicate that fine-tuned models with LoTRA outperform or perform comparably with full fine-tuning models and models trained from scratch. 
Moreover, KANs consistently outperform MLPs in solving Allen-Cahn equations, showing their potential in science-related tasks.

\begin{figure*}[!t]
    \centering
    \begin{minipage}[t]{0.3\textwidth} 
        \centering
    \subfigure[$\epsilon=1.0$, $\text{cr}=1/4$]{%
    \includegraphics[width=\textwidth]{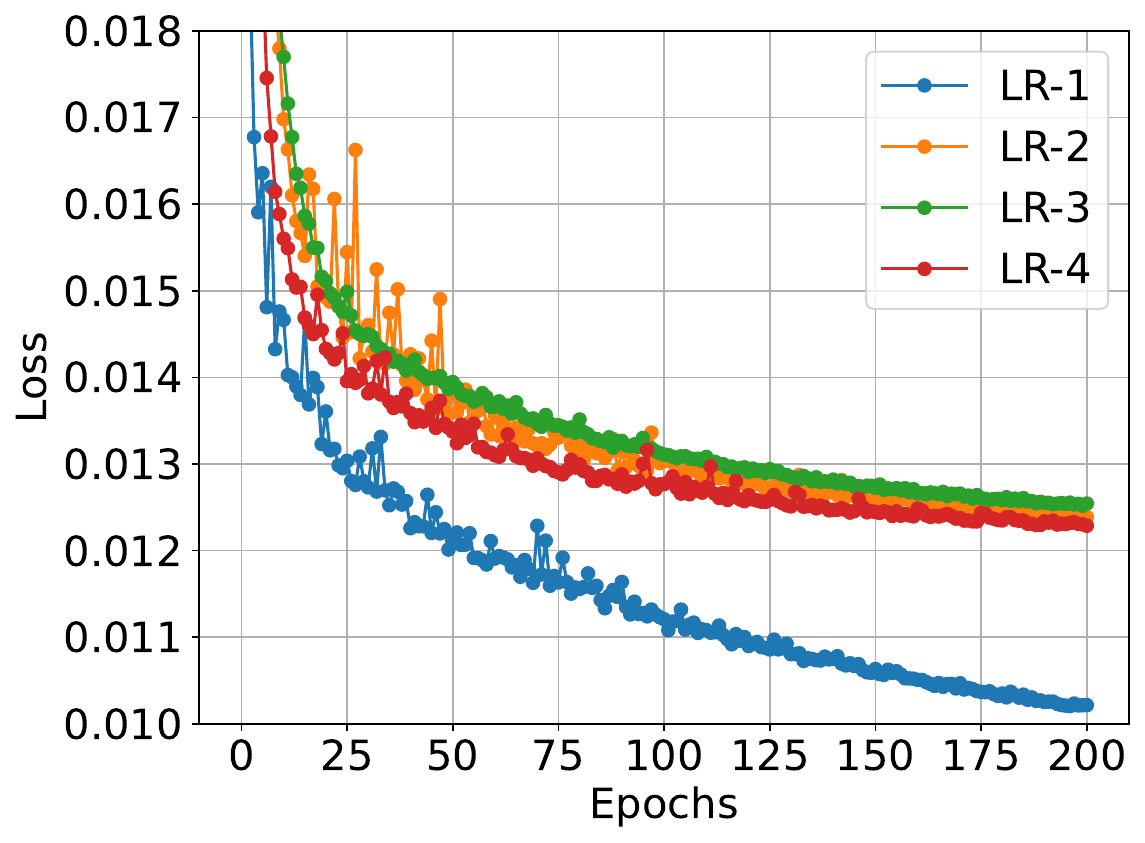} \label{fig_cheby_legendre_eps1.0_cr1/4_allen_cahn}}
    \caption{Fine-tuning trajectories of Chebyshev KANs using LoTRA under four strategies of learning rate selection (denoted as ``LR-1" to ``LR-4") for solving Allen-Cahn equations.}
    \label{fig_cheby_legendre_lr_allen_cahn}
    \end{minipage}
    \hfill
    \begin{minipage}[t]{0.66\textwidth} 
        \centering
    \subfigure[$\epsilon=0.1$]{%
        \includegraphics[width=0.48\textwidth]{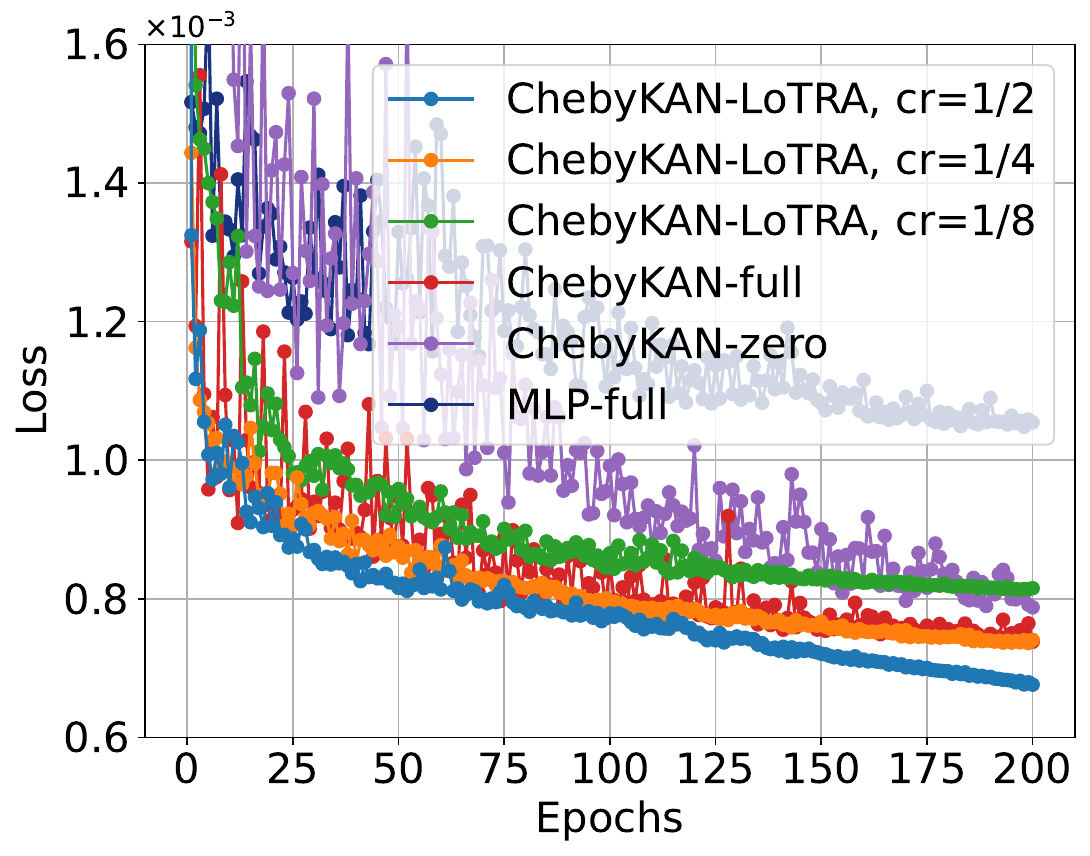} \label{fig_cheby_legendre_eps0.1_allen_cahn}}
    \subfigure[$\epsilon=1.0$]{%
        \includegraphics[width=0.48\textwidth]{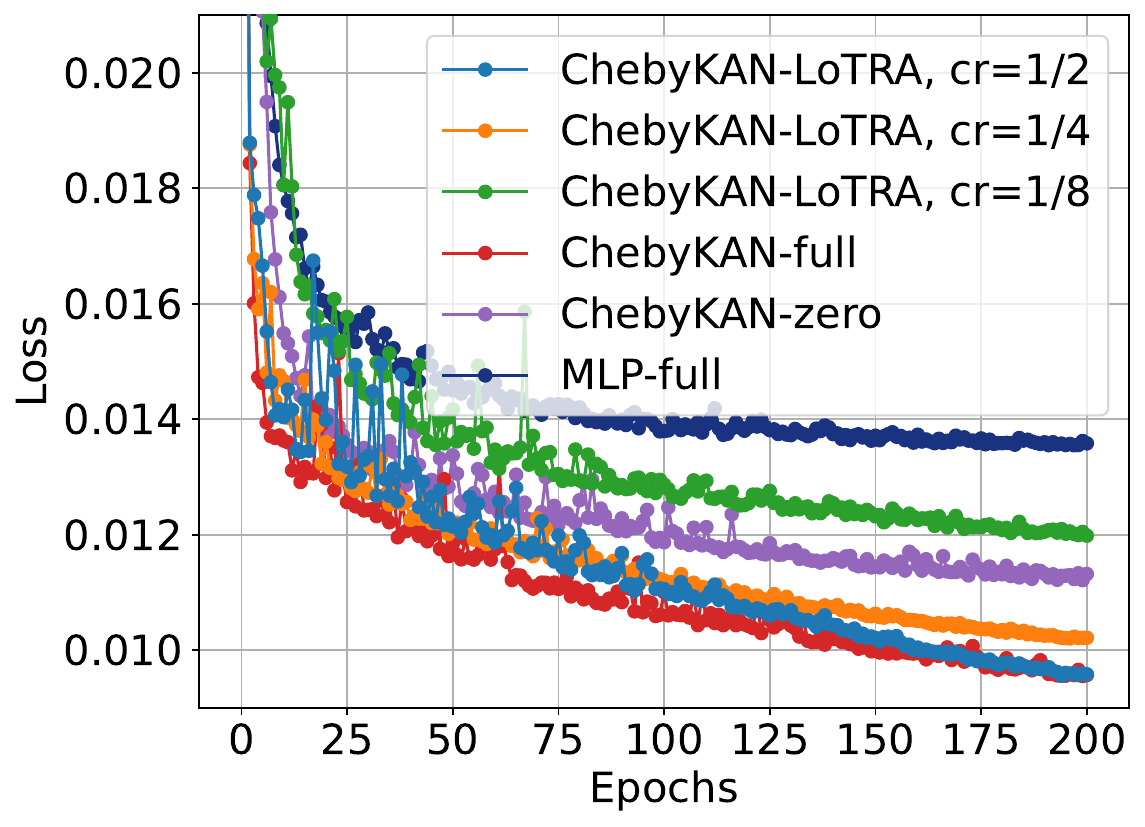} \label{fig_cheby_legendre_eps1.0_allen_cahn}}
    \caption{Fine-tuning trajectories of Chebyshev KANs using LoTRA, compared to fully updated KANs and MLPs, for solving Allen-Cahn equations.}
    \label{fig_cheby_legendre_allen_cahn}
    \end{minipage}
\end{figure*}

\subsubsection{Hyperbolic Equations}

We consider a class of hyperbolic equations with varying parameters $\lambda \in \mathbb{R}$:
\begin{align}
        \frac{\partial^2 u(t,\bm{x};\lambda)}{\partial t^2} - \Delta u(t,\bm{x};\lambda) & = f(t,\bm{x};\lambda), \quad (t,\bm{x}) \in [0,1] \times \Omega, \nonumber\\
        u(t,\bm{x};\lambda) & = g(t,\bm{x};\lambda), \quad (t,\bm{x}) \in [0,1] \times \partial \Omega,\nonumber\\
        u(0,\bm{x};\lambda) & = h(\bm{x};\lambda), \quad \bm{x} \in \Omega,\nonumber\\
        \frac{u(0,\bm{x};\lambda)}{\partial t} & = \Bar{h}(\bm{x};\lambda), \quad \bm{x} \in \Omega,
\end{align}
where the temporal and the spatial domains are defined as $[0,1]$ and $\Omega = \{\bx \in \mathbb{R}^{2}: \left\|\bx \right\|_2 \leq 1\}$, respectively. The exact solution $\bu(\bx;\lambda)$ with the parameter $\lambda$ is defined as $u(\bx;\lambda) = \big(e^{t^2}-1\big) \sin \big(\frac{\pi}{2}\big( 1 - \left\|\bm{x} \right\|_2 \big)^{2.5} \big) + \lambda \cdot \big(e^{t^2}-1\big) \sin \big(\frac{\pi}{2}\big( 1 - \left\|\bm{x} \right\|_2\big ) \big)$. 
We first pre-train a KAN model on the PDE with $\lambda =0$, and then fine-tune it on PDEs with $\lambda=0.1$ and $\lambda = 1$.

We conduct similar experiments as in Sections \ref{sec_transfer_elliptic} and \ref{sec_transfer_allen_cahn} for solving hyperbolic equations using KANs with LoTRA, adopting four different learning rate strategies to verify the theoretical results in \Cref{theorem_efficient}. 
The four strategies follow those in Sections  \ref{sec_transfer_elliptic} and \ref{sec_transfer_allen_cahn}, where the first strategy is consistent with the theorem, while the other three deviate from the theoretically optimal learning rate selection.   
The fine-tuning trajectories of KANs with LoTRA under these learning rates strategies are illustrated in Figures S9 and S11 in the supplementary material. 
Across different transfer learning tasks, including varying parameter values $\lambda$, different compression ratios, and various KAN variants, we observe that the first strategy, derived from our theoretical results, consistently achieves the lowest training loss in most cases.
In some instances, the first strategy may exhibit slightly lower performance. This is primarily due to our naive assumption of setting the constant in $\Theta(\cdot)$ to be 1e-2 for all learning rates, which is not the best value. However, given the large number of variables involved, we must fix the constant term for practical feasibility and comparison.
This empirical observation further validates \Cref{theorem_efficient}, demonstrating the effectiveness of the proposed learning rate selection strategy. 
The superiority of the first learning rate strategy are even more evident in hyperbolic equations compared to those observed in elliptic and Allen-Cahn equations. This is mainly due to the higher complexity of hyperbolic equations, which involve second-order derivatives on the temporal variable, making them more challenging to solve.  
Therefore, the benefits of the theoretically guided learning rate selection are more apparent when applied to hyperbolic equations than to elliptic and Allen-Cahn equations.

We compare the fine-tuned KANs using the proposed LoTRA with several baseline methods for solving hyperbolic equations. The baseline models are the same as those examined in Sections \ref{sec_transfer_elliptic} and \ref{sec_transfer_allen_cahn}. The visualized results are presented in Figures S10 and S12 in the supplementary material. 
The results demonstrate that KAN models with LoTRA and $\text{cr}=1/2$ outperform the vanilla fine-tuning method with full parameter updates. Additionally, they significantly outperform models trained from scratch without transfer learning.  
We also observe that LoTRA with larger $\text{cr}$ (i.e., larger core tensors) achieve lower training loss due to increased expressiveness.
This observation differs slightly from the results for elliptic equations but is consistent with those for Allen-Cahn equations.
Notably, the performance gap in hyperbolic equations becomes even more obvious and amplified across different core tensor sizes. This is primarily because hyperbolic equations, which involve higher-order derivatives in the temporal domain, are inherently more challenging and demand higher model expressiveness.  
To further evaluate performance, we compute the relative error of the fine-tuned models, as shown in Tables SIII and SIV in the supplementary material. The results indicate that fine-tuned models with LoTRA outperform or perform comparably with full fine-tuning models and models trained from scratch. 
Moreover, KANs consistently and significantly outperform MLPs in solving hyperbolic equations, further showing their potential in science-related tasks. 
Both LoTRA-based fine-tuning and full parameter updates outperform models trained from scratch in most cases, suggesting that transfer learning effectively captures shared information from the pre-training task, leading to faster convergence.

\subsection{Slim KANs}
In the previous section, we demonstrated the effectiveness of LoTRA in transfer learning for KANs, successfully solving three classes of PDEs with faster convergence and significantly less parameter size. 
In this subsection, we further evaluate slim KANs, developed based on the LoTRA framework without pre-trained parameters, on function representation and image classification tasks.

\subsubsection{Trigonometric Function}

We evaluate the expressiveness of slim KANs on a simple function representation task. Specifically, we aim to represent a two-dimensional trigonometric function given by $u(x,y) = \sin(\pi x) \sin(\pi y)$. 
To introduce complexity, we add Gaussian noise with zero mean and a standard deviation of $0.05$. 
We experiment with various compression ratios (core tensor size) and basis functions.
For this task, we use three-layer KANs with a hidden dimension of  $n_{\ell} = 32$ and $n_d = 8$. Given the small input and output dimensions, we only apply compression to the hidden layers.
We use the MLP with a similar parameter size (hidden dimension to be $128$) as the baseline model to compare its performance with KANs on the trigonometric function representation task.

\begin{figure}[tb!]
    \centering
    \subfigure[Training Loss]{%
        \includegraphics[width=0.23\textwidth]{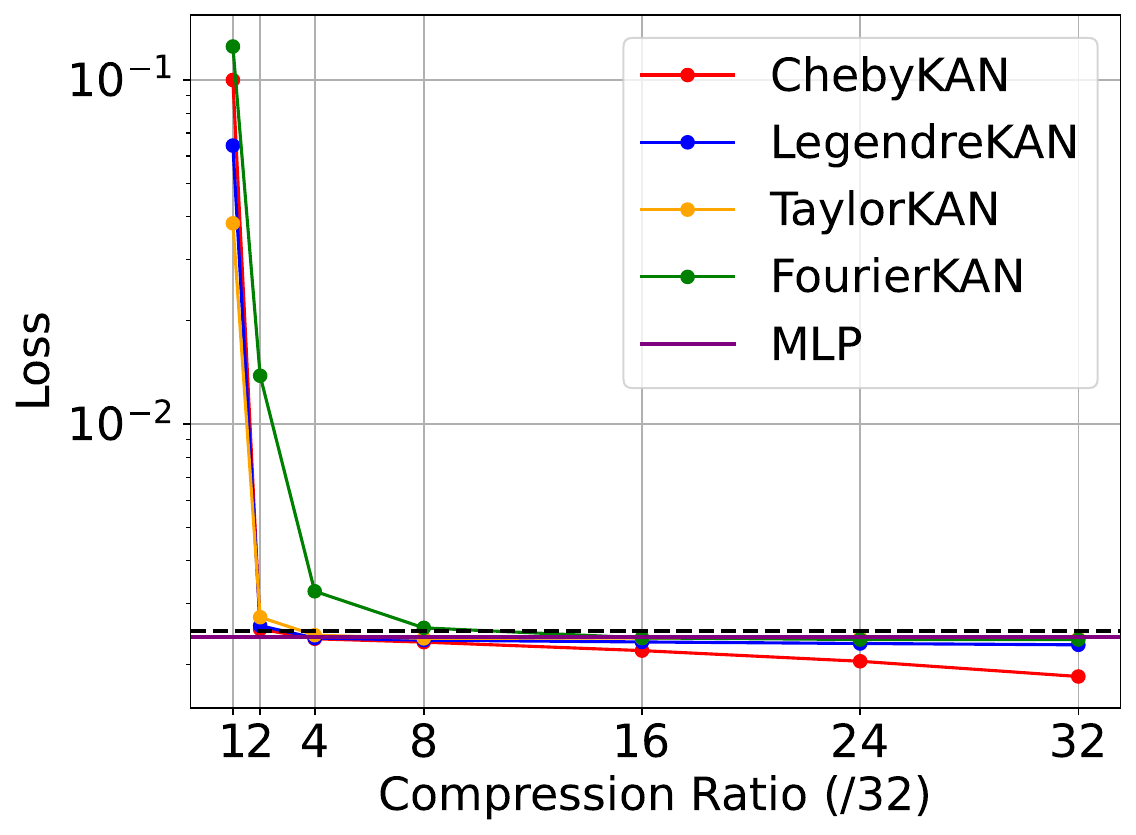} \label{fig_slim_kan_sin_loss}}
    \subfigure[Validation Error]{%
        \includegraphics[width=0.23\textwidth]{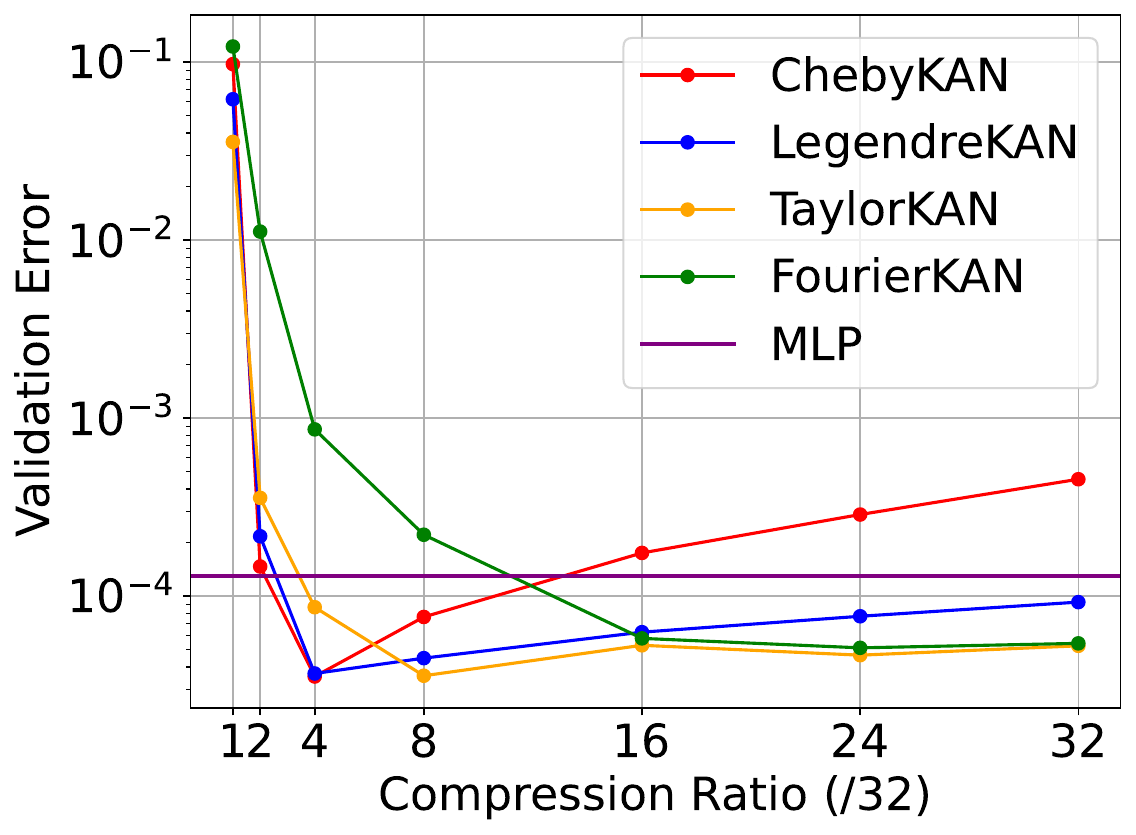} \label{fig_slim_kan_sin_error}}
    \caption{Relationship between the compression ratio and the performances on representing a trigonometric function. (a) Training Loss: The training loss decreases with an increasing compression ratio. (b) Validation Error: The validation error shows a U-shape, indicating a balance between model complexity and generalization.}
    \label{fig_slim_kan_sin}
\end{figure}

\begin{figure}[tb!]
    \centering
    \subfigure[cr is 4/32]{%
        \includegraphics[width=0.15\textwidth]{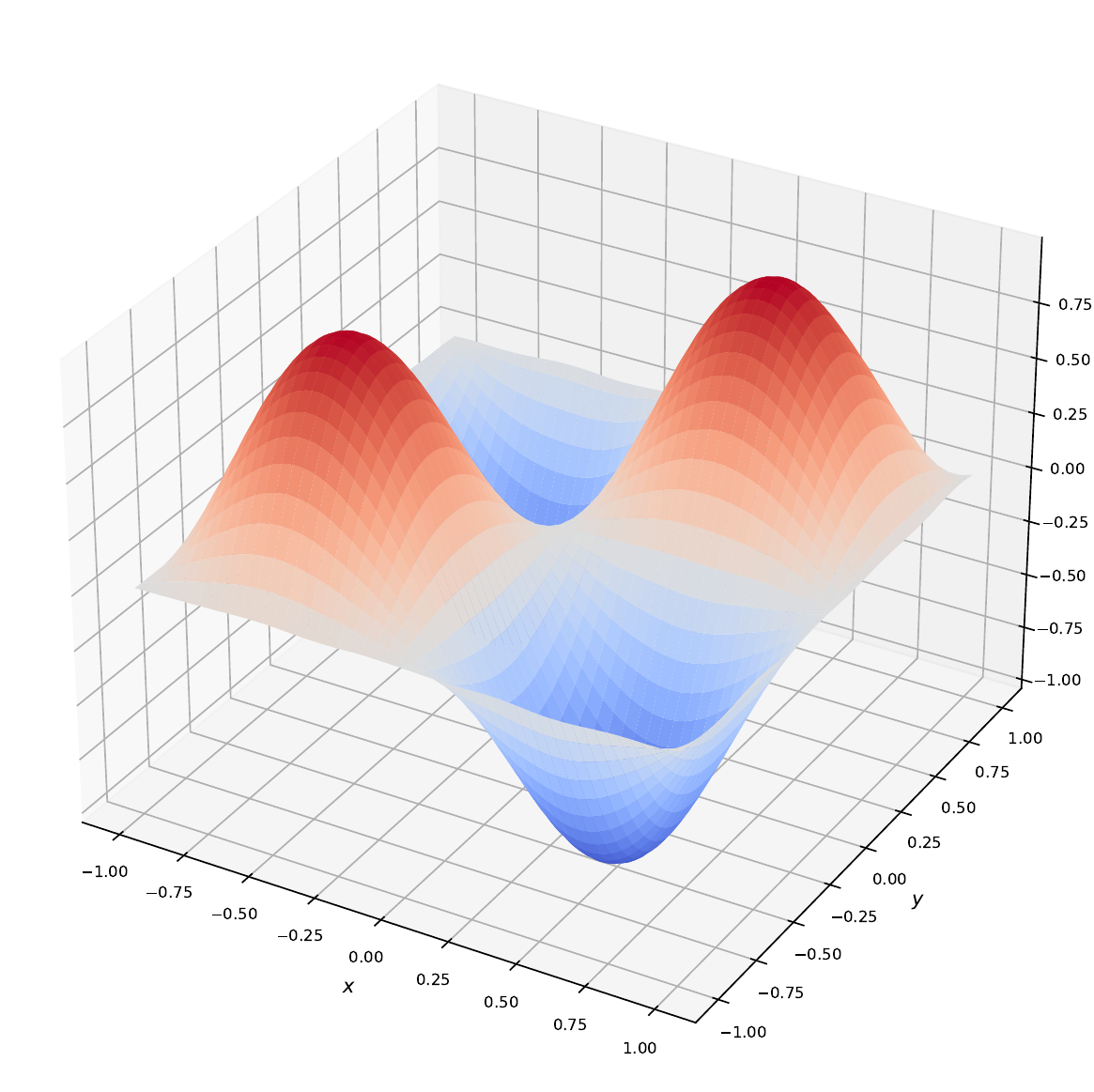} \label{fig_slim_kan_sin_cr_4}}
    \subfigure[cr is 16/32]{%
        \includegraphics[width=0.15\textwidth]{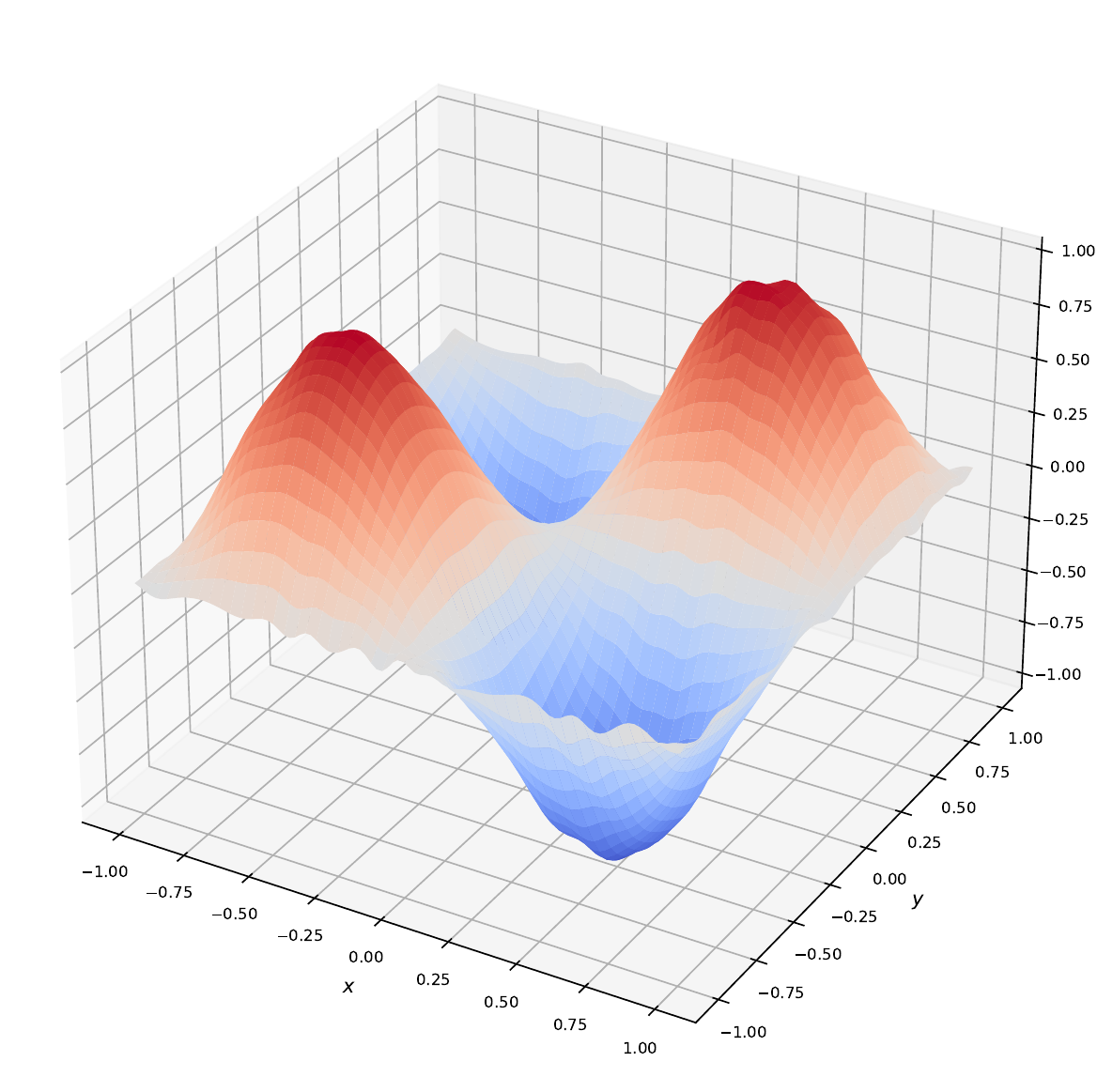} \label{fig_slim_kan_sin_cr_16}}
    \subfigure[cr is 32/32]{%
    \includegraphics[width=0.15\textwidth]{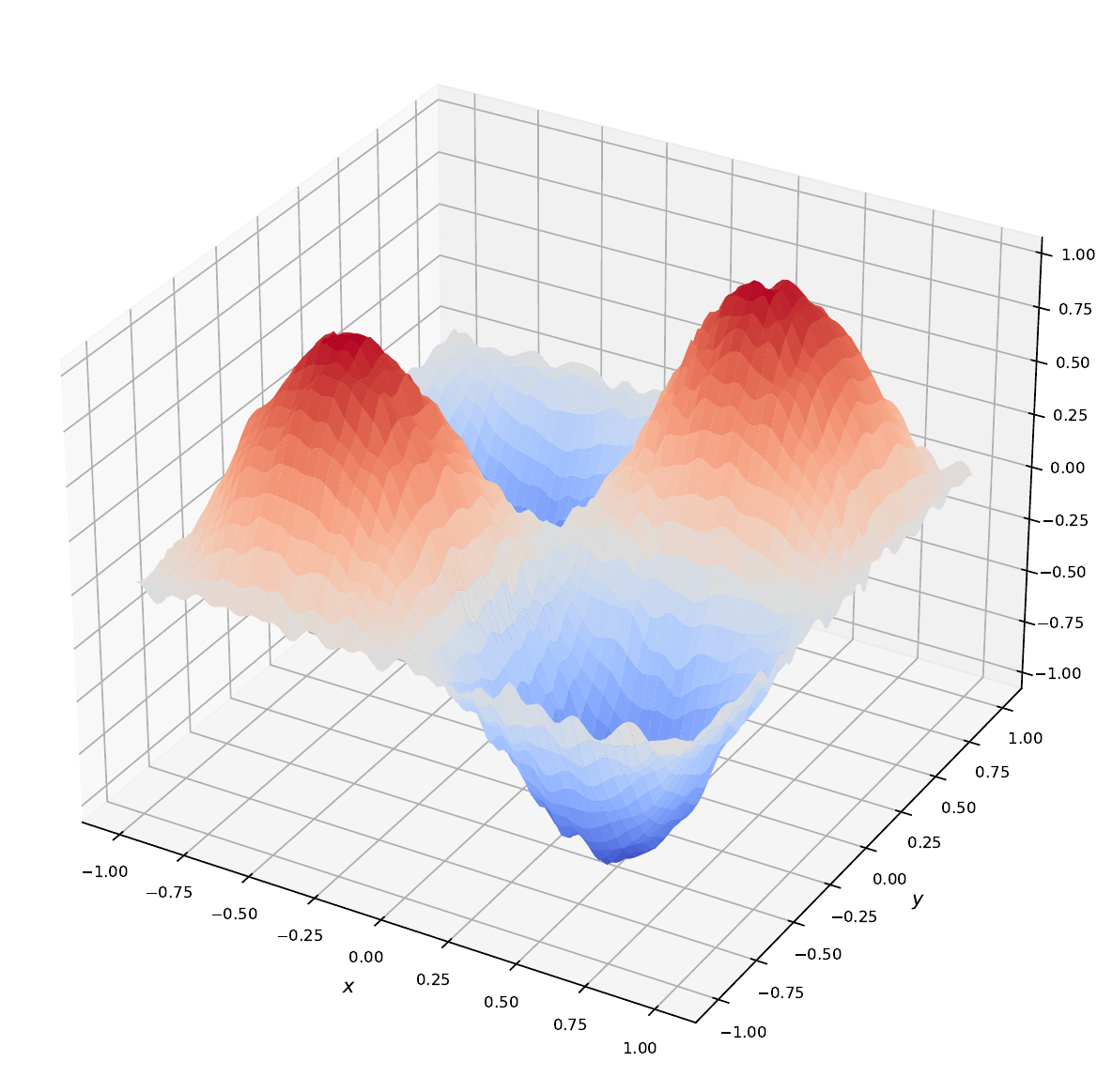} \label{fig_slim_kan_sin_cr_32}}
    \caption{Visualization of the generated function by slim KAN models with different compression ratios for representing a trigonometric function. The figures illustrate that models with a larger compression ratio tend to overfit the noise, producing less smooth results. In contrast, models with smaller compression ratios generate smoother functions, reflecting better generalization.}
    \label{fig_slim_kan_sin_cr_vis}
\end{figure}

The training loss with respect to the compression ratio (cr) is visualized in Figure \ref{fig_slim_kan_sin}.
From the results, we observe that the training loss decreases monotonically as the compression ratio increases, highlighting the improved expressiveness of slim KANs with larger core tensors. This observation is consistent with our intuition. However, when considering the validation error, the curve exhibits a U-shape, indicating overfitting in slim KANs with larger core tensor sizes, which is also within our expectations.
Here, in the simple example, we observe that the MLP demonstrates comparable performance to KANs.
Furthermore, the dashed line in the figure represents the expected training loss due to the introduced noise. 
Interestingly, slim KANs with larger core tensors achieve training losses smaller than this threshold, implying that they are fitting the noise. This phenomenon is further corroborated by the observed increase in validation error when the model achieves lower training losses than the noise level.
The visualization of generated functions is shown in \Cref{fig_slim_kan_sin_cr_vis}. Models with larger compression ratios tend to overfit the noise, resulting in a nonsmooth surface for the generated shape.
In summary, our observations confirm that Slim KANs are highly expressive for function representation tasks. Additionally, the low-rank structure introduced by LoTRA acts as an effective regularization mechanism to mitigate overfitting.

\subsubsection{Nonsmooth and Sharp Function}

We evaluate the performance of slim KANs on a more challenging task of representing a two-dimensional nonsmooth and sharp function. Due to the growing complexity of the function, we use a larger KAN model with $n_{\ell}=128$ hidden dimensions and $n_d=8$ basis functions. As a baseline for comparison, we use an MLP with a hidden dimension of $1024$.

\begin{figure}[tb!]
    \centering
    \subfigure[Training Loss]{%
        \includegraphics[width=0.23\textwidth]{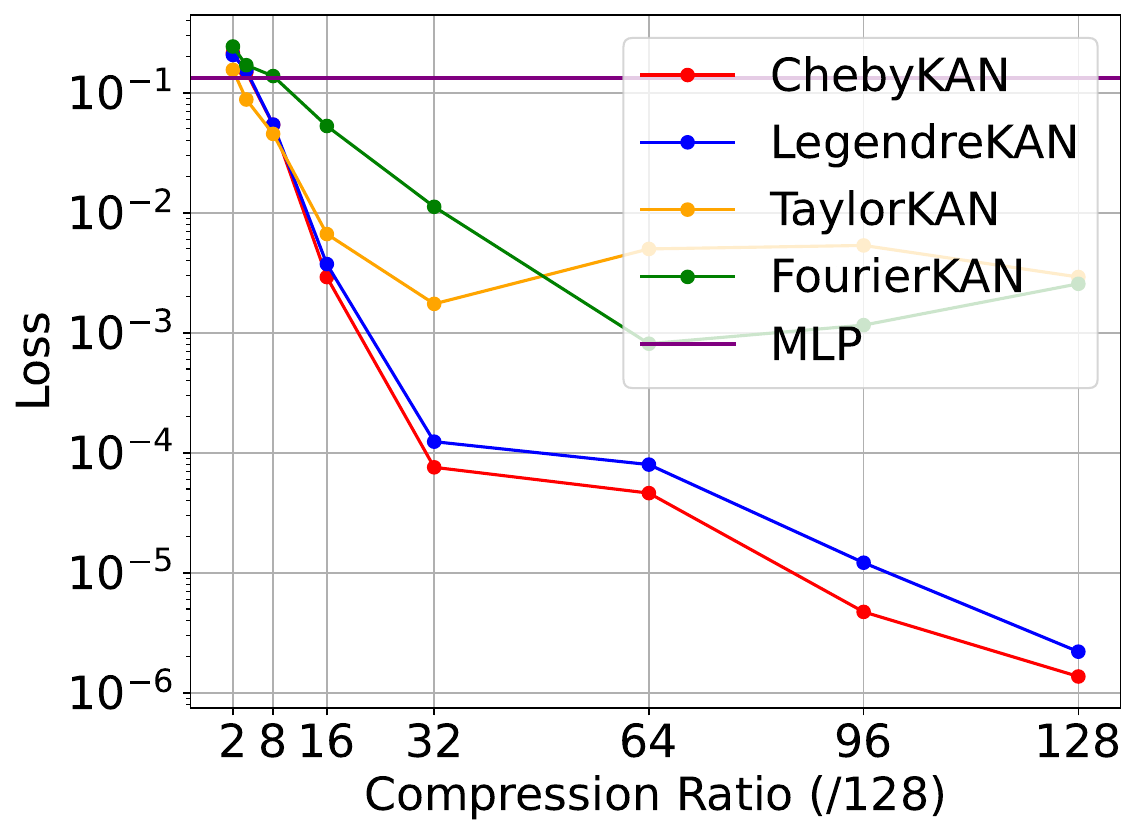} \label{fig_slim_kan_nonsmooth_loss}}
    \subfigure[Validation Error]{%
        \includegraphics[width=0.23\textwidth]{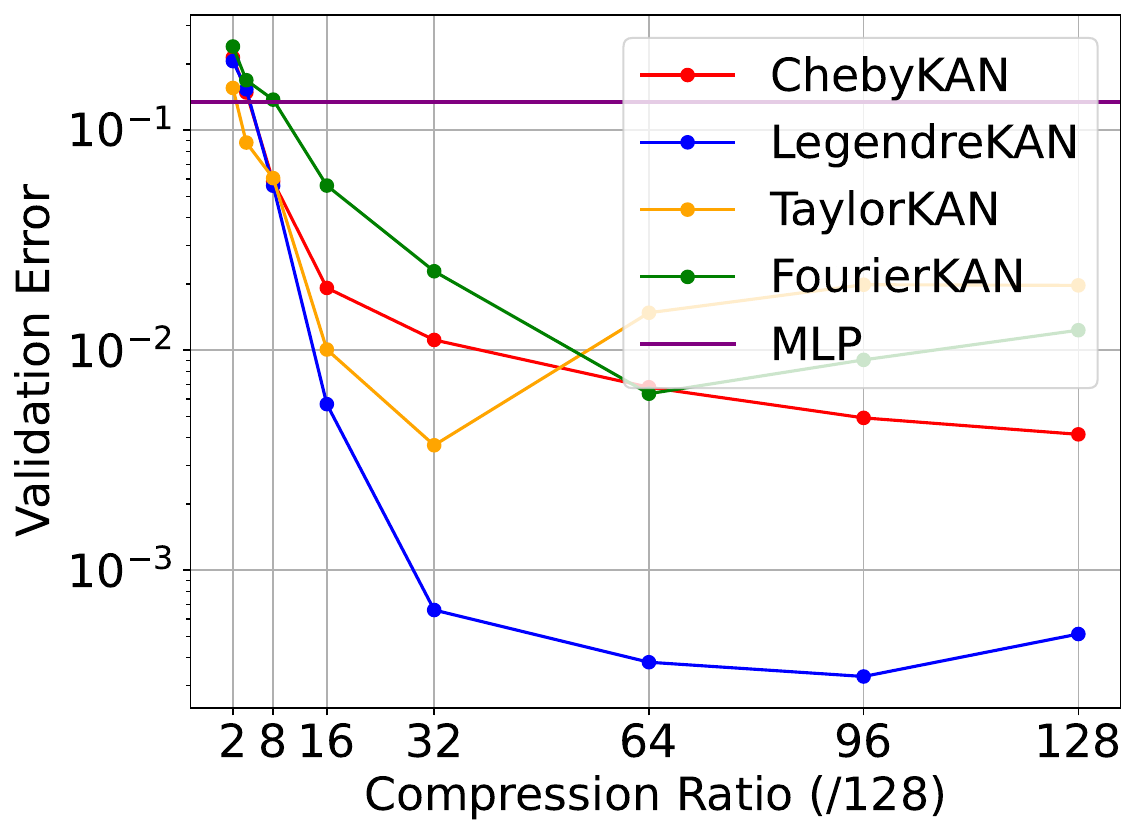} \label{fig_slim_kan_nonsmooth_error}}
    \caption{Relationship between the compression ratio and the performances on representing a nonsmooth function. (a) Training Loss: The training loss decreases with an increasing compression ratio. (b) Validation Error: The validation error shows a similar trend, indicating higher expressiveness with a larger compression ratio.}
    \label{fig_slim_kan_nonsmooth}
\end{figure}

\begin{figure}[tb!]
    \centering
    \subfigure[cr is 4/128]{%
        \includegraphics[trim=0 0 0 20, clip, width=0.15\textwidth]{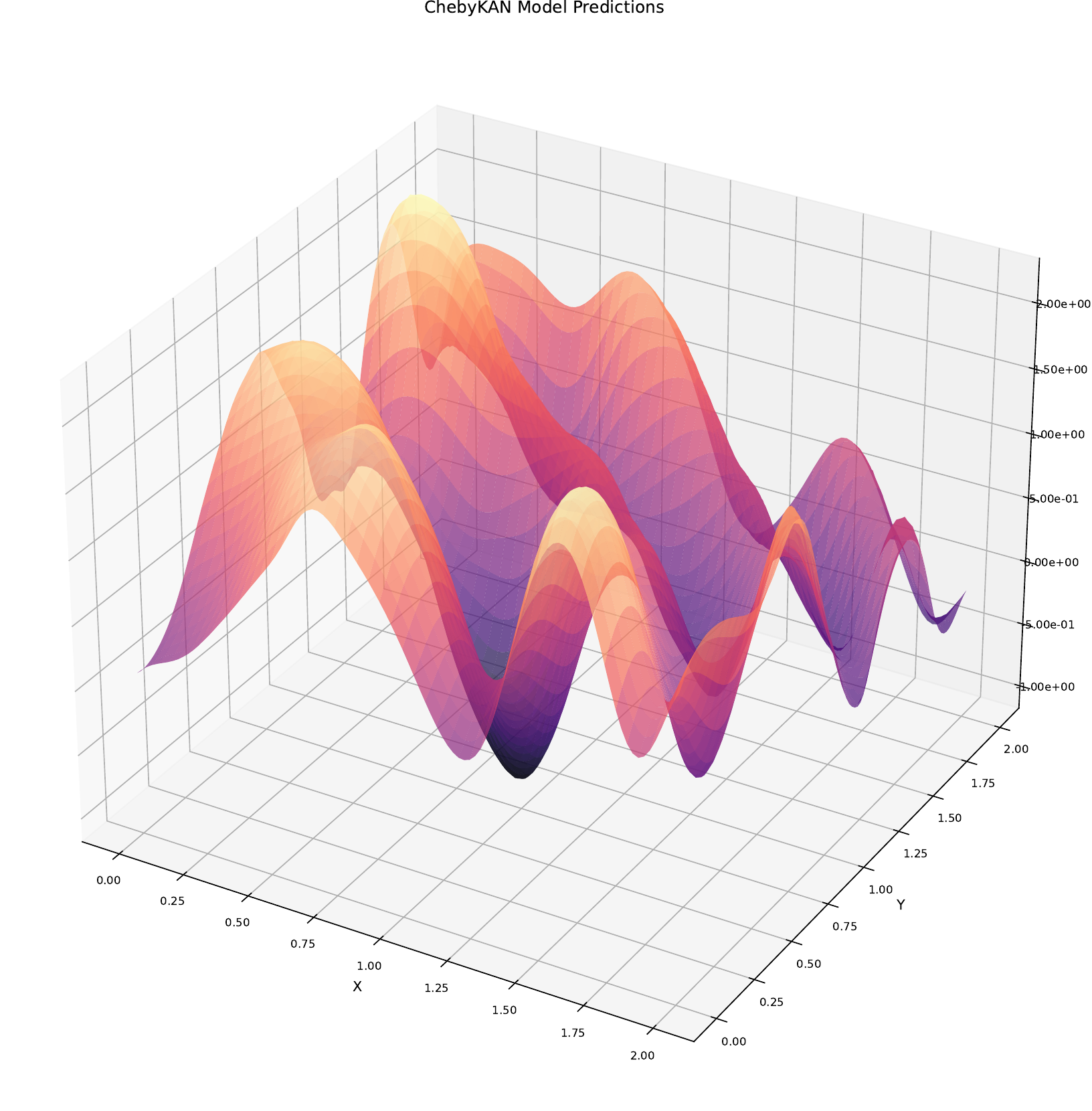} \label{fig_slim_kan_nonsmooth_cr_4}}
    \subfigure[cr is 64/128]{%
        \includegraphics[trim=0 0 0 20, clip,width=0.15\textwidth]{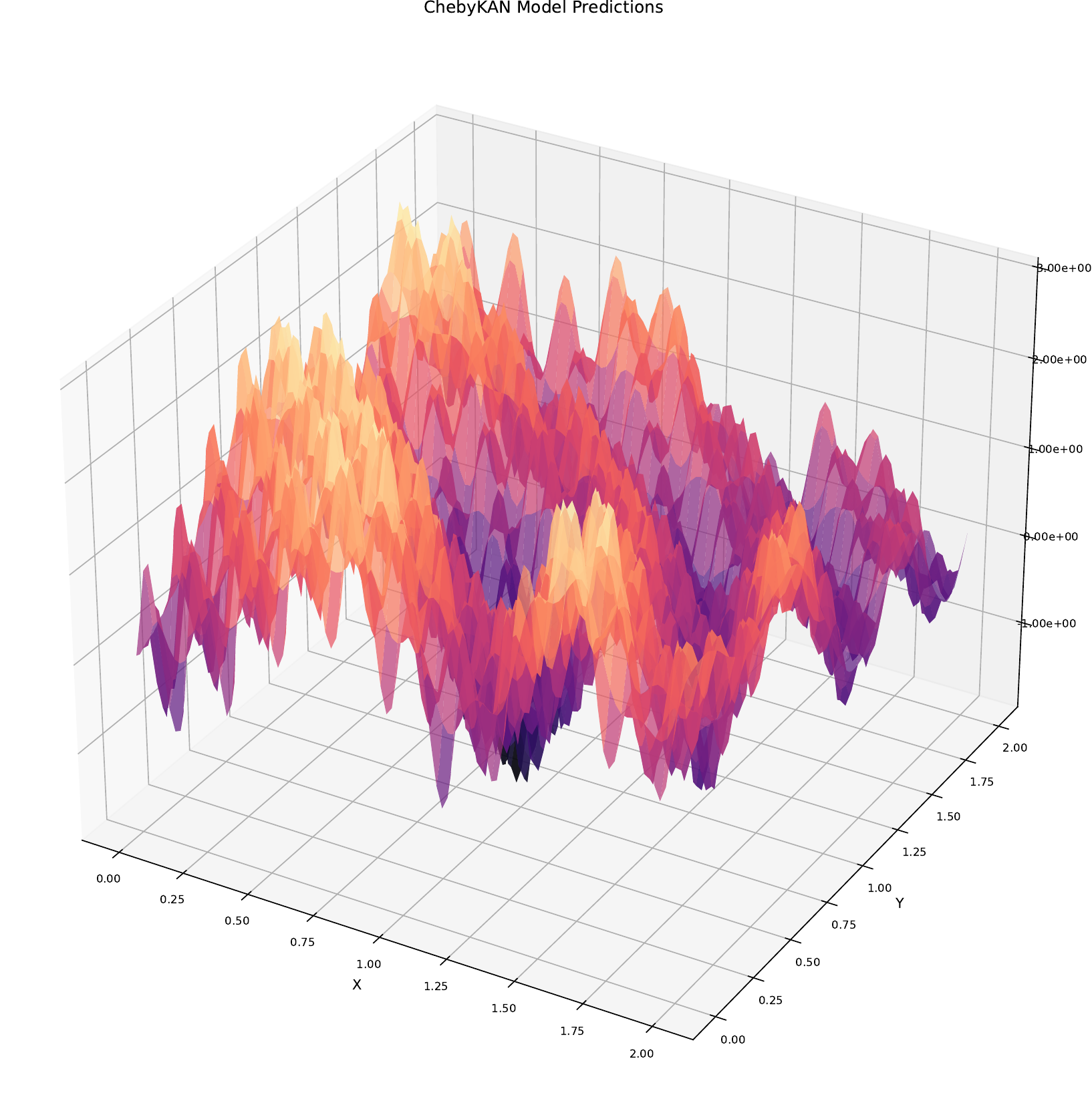} \label{fig_slim_kan_nonsmooth_cr_64}}
    \subfigure[cr is 128/128]{%
    \includegraphics[trim=0 0 0 20, clip,width=0.15\textwidth]{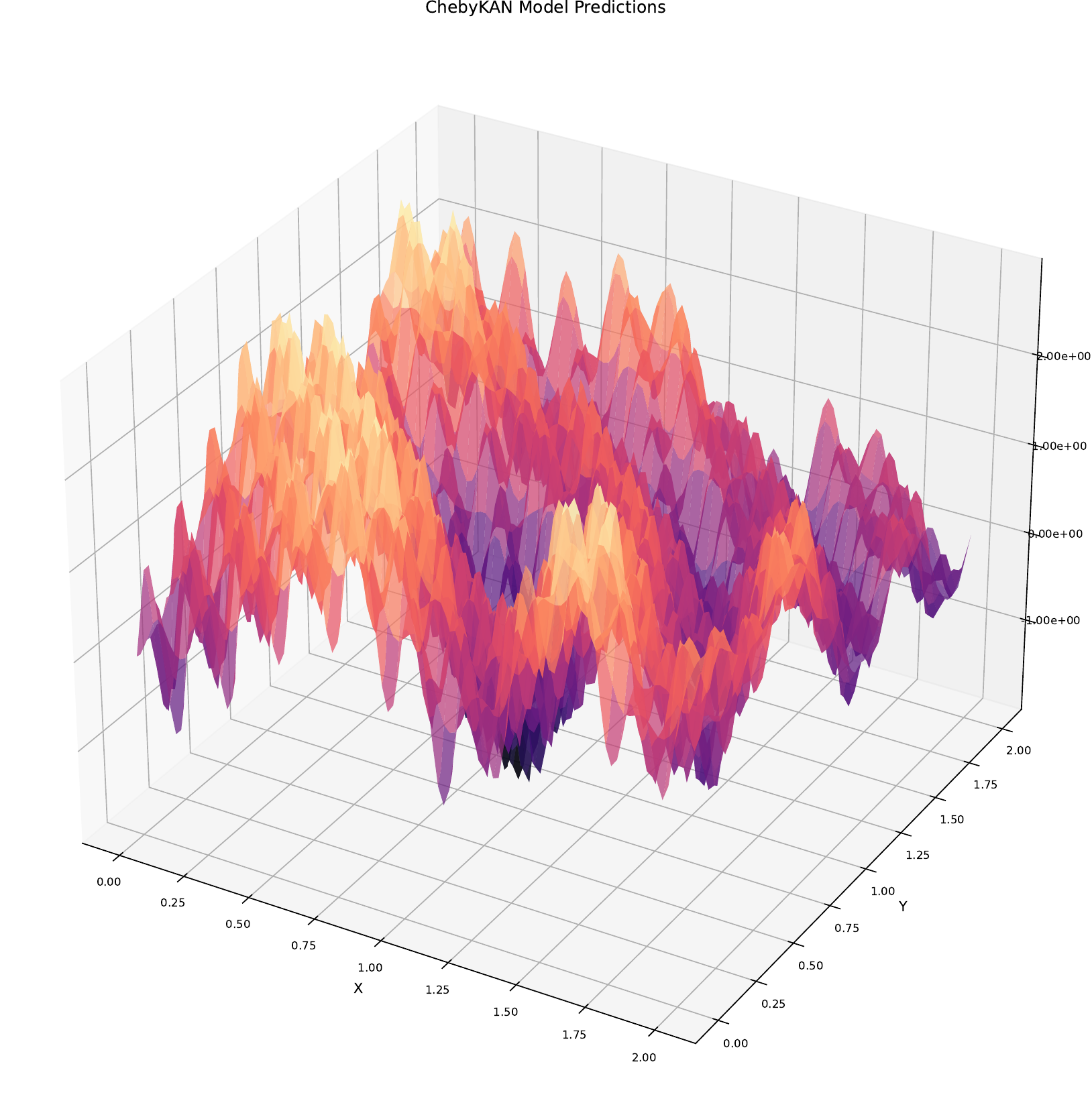} \label{fig_slim_kan_nonsmooth_cr_128}}
    \caption{Visualization of the generated function by KAN models with different compression ratios for representing a nonsmooth function. The figures highlight that models with larger compression ratios exhibit higher expressiveness.}
    \label{fig_slim_kan_nonsmooth_cr_vis}
\end{figure}

The relationship between the compression ratio (core tensor size) and model performance is visualized in \Cref{fig_slim_kan_nonsmooth}. The figure demonstrates that larger compression ratios correspond to higher model complexity and expressiveness, leading to lower training loss and validation error. This behavior differs slightly from the results for the trigonometric function, as the trigonometric function is relatively simple, whereas the nonsmooth function is significantly more challenging. Even with an enlarged model size, the KAN model does not overfit the nonsmooth function, and the characteristic U-shape observed in the trigonometric function task does not appear. Additionally, the MLP fails to accurately represent the nonsmooth function, on the contrary, KANs achieve substantially lower training loss and validation error, showing the superior ability of KANs in function representation compared to MLP. 
The visualization of the generated functions is shown in \Cref{fig_slim_kan_nonsmooth_cr_vis}. The model with $\text{cr}=4/64$ fails to capture the oscillatory trajectories of the function due to its limited expressiveness. In contrast, the slim KAN model with a larger core tensor size effectively captures more intricate details of this complicated nonsmooth function.

\subsubsection{Image Classification: MNIST and CIFAR-10 Datasets}

We further apply Slim KANs to image classification tasks on the MNIST and CIFAR-10 datasets. For the MNIST dataset, we adopt the three-layer feed-forward Slim KANs model, where the input and hidden layers are compressed, while the output layer remains uncompressed. The input dimension, hidden dimension, and output dimension are in the size of $28 \times 28$, $128$, and $10$, respectively. Following the experimental settings in~\cite{ss2024chebyshev}, the number of basis functions is set to $n_d = 4$. For the CIFAR-10 dataset, which is more complex, we use a one-layer ResNet~\cite{he2016deep} consisting of three convolutional layers, with the hidden feed-forward and the output layer replaced by slim KAN layers. The hidden dimension of the slim KAN layer is set to $n_{\ell}=128$ and the number of basis functions is $n_d = 4$. After the convolutional layers, the input to the slim KAN layer has a dimension of 256. Compression is applied only to this layer, while the output layer remains as a standard KAN layer. Since $n_d$ is much smaller than the hidden dimension, we do not compress the third mode of the tensor, meaning the third mode of the core tensor matches the original tensor. 
We also compare slim KANs against MLPs as baseline models with a comparable parameter size (denoted as ``MLP-small") and significantly larger parameter size (denoted as ``MLP-large", with approximately four times more parameters).
The performance of Slim KANs with varying compression ratios and different basis functions is summarized in Tables \ref{tab_minist} and \ref{tab_cifar10}.

We observe from \Cref{tab_minist} that models with larger compression ratios (i.e., larger core tensors) demonstrate greater expressiveness and achieve higher accuracy. However, in this example, slim KANs and original KANs perform slightly worse than MLPs of similar parameter sizes. A similar observation was reported in \cite{yu2024kan}, which conducted a comprehensive comparison between KANs and MLPs across various tasks. The study found that while KANs often outperform MLPs in function representation tasks, they tend to slightly underperform in computer vision and machine learning tasks. Our results are consistent with these observations.
Despite this, we emphasize that slim KANs, with significantly fewer parameters, maintain competitive and satisfying performance, demonstrating their potential for parameter-efficient tasks. This emphasizes the potential and promise of slim KANs in applications where reducing parameter sizes is crucial without significantly compromising accuracy.

\Cref{tab_cifar10} further highlights the expressiveness and potential of slim KANs models when integrated with ResNet. Similarly, we observe that a larger compression ratio enhances expressiveness and leads to higher accuracy. Notably, when combined with a one-layer ResNet, the model outperforms the baseline MLP models with comparable parameter sizes. These results demonstrate the potential of slim KANs when integrated with advanced architectures in specific applications.

\begin{table}[htb!]
\footnotesize
\begin{tabular}{|ll|c|ll|c|}
\hline
\multicolumn{2}{|l|}{Method}                                      & Accuracy& \multicolumn{2}{l|}{Method}                                     & Accuracy \\ \cline{1-2} \cline{4-5}
\multicolumn{1}{|l|}{Type}                          & cr (/128) &  (\%)                         & \multicolumn{1}{l|}{Type}                         & cr (/128) & (\%)                          \\ \hline
\multicolumn{1}{|l|}{\multirow{6}{*}{Chebyshev}}    &  8  & 94.48 & \multicolumn{1}{l|}{\multirow{7}{*}{Taylor}}    & 8 & 95.15 \\ \cline{2-3} \cline{5-6} 
\multicolumn{1}{|l|}{} & 16 & 96.21 & \multicolumn{1}{l|}{}    & 16 & 96.77 \\ \cline{2-3} \cline{5-6} 
\multicolumn{1}{|l|}{} & 32 & 96.59 & \multicolumn{1}{l|}{} & 32 & 96.86    \\ \cline{2-3} \cline{5-6} 
\multicolumn{1}{|l|}{}  &  64 & 96.72 & \multicolumn{1}{l|}{}  & 64 & 96.64  \\ \cline{2-3} \cline{5-6} 
\multicolumn{1}{|l|}{} & 96 & 96.57 & \multicolumn{1}{l|}{}  &  96 & 96.36  \\ \cline{2-3} \cline{5-6} 
\multicolumn{1}{|l|}{} & 128 & 97.18  & \multicolumn{1}{l|}{}  & 128  &  96.68  \\ \hline
\multicolumn{1}{|l|}{\multirow{7}{*}{Legendre}} &  8  & 95.05 & \multicolumn{1}{l|}{\multirow{6}{*}{Fourier}} & 8  &  95.01   \\ \cline{2-3} \cline{5-6} 
\multicolumn{1}{|l|}{} & 16 & 95.79 & \multicolumn{1}{l|}{}                             & 16  & 96.20 \\ \cline{2-3} \cline{5-6} 
\multicolumn{1}{|l|}{} & 32 & 96.45  & \multicolumn{1}{l|}{} &  32 & 96.48  \\ \cline{2-3} \cline{5-6} 
\multicolumn{1}{|l|}{} & 64 &  96.76 & \multicolumn{1}{l|}{}  & 64 & 96.81 \\ \cline{2-3} \cline{5-6} 
\multicolumn{1}{|l|}{} &  96 & 97.42 & \multicolumn{1}{l|}{}  & 96 & 97.54 \\\cline{2-3} \cline{5-6} 
\multicolumn{1}{|l|}{} & 128 & 97.73 & \multicolumn{1}{l|}{} & 128 & 97.57 \\ \hline
\multicolumn{2}{|l|}{MLP-Small} & 98.08 & \multicolumn{2}{l|}{MLP-Large} & 98.47 \\ \hline
\end{tabular}
\caption{Accuracy of slim KANs, MLP with comparable parameters (MLP-small), and MLP with an enlarged model size (MLP-large) for the image classification task on the MNIST dataset.}
\label{tab_minist}
\end{table}

\begin{table}[htb!]
\begin{tabular}{|ll|c|ll|c|}
\hline
\multicolumn{2}{|l|}{Method}                                      & Accuracy & \multicolumn{2}{l|}{Method}                                     & Accuracy \\ \cline{1-2} \cline{4-5}
\multicolumn{1}{|l|}{Type}                          & cr (/128) &       (\%)                             & \multicolumn{1}{l|}{Type}                         & cr (/128) &     (\%)                               \\ \hline
\multicolumn{1}{|l|}{\multirow{7}{*}{Chebyshev}} & 8 &  69.67 & \multicolumn{1}{l|}{\multirow{7}{*}{Taylor}} & 8 & 69.83  \\ \cline{2-3} \cline{5-6} 
\multicolumn{1}{|l|}{} & 16 & 73.43 & \multicolumn{1}{l|}{} &  16  & 73.33  \\ \cline{2-3} \cline{5-6} 
\multicolumn{1}{|l|}{} &  32  & 74.92  & \multicolumn{1}{l|}{} & 32 & 75.33  \\ \cline{2-3} \cline{5-6} 
\multicolumn{1}{|l|}{} &  48 & 75.92  & \multicolumn{1}{l|}{} & 48 & 75.82 \\ \cline{2-3} \cline{5-6} 
\multicolumn{1}{|l|}{} &  64  & 76.24 & \multicolumn{1}{l|}{} & 64 & 76.10 \\ \cline{2-3} \cline{5-6} 
\multicolumn{1}{|l|}{} &  96 &  76.50 & \multicolumn{1}{l|}{} & 96 & 76.24 \\ \cline{2-3} \cline{5-6} 
\multicolumn{1}{|l|}{}                              &  128 & 76.63  & \multicolumn{1}{l|}{} & 128 & 76.27 \\ \hline
\multicolumn{1}{|l|}{\multirow{7}{*}{Legendre}} & 8 &  69.88  & \multicolumn{1}{l|}{\multirow{7}{*}{Fourier}} &  8 & 68.95 \\ \cline{2-3} \cline{5-6} 
\multicolumn{1}{|l|}{} & 16 &  73.48 & \multicolumn{1}{l|}{} & 16 & 72.96  \\ \cline{2-3} \cline{5-6} 
\multicolumn{1}{|l|}{}  & 32 & 75.44 & \multicolumn{1}{l|}{}  &  32 &  74.41   \\ \cline{2-3} \cline{5-6} 
\multicolumn{1}{|l|}{} & 48 & 75.55  & \multicolumn{1}{l|}{}  &  48 & 75.49 \\ \cline{2-3} \cline{5-6} 
\multicolumn{1}{|l|}{} & 64 & 75.39  & \multicolumn{1}{l|}{} &  64 & 75.63  \\ \cline{2-3} \cline{5-6} 
\multicolumn{1}{|l|}{}  &  96 & 75.62  & \multicolumn{1}{l|}{} & 96 &  76.50  \\ \cline{2-3} \cline{5-6} 
\multicolumn{1}{|l|}{} & 128 & 76.13 & \multicolumn{1}{l|}{} & 128 & 76.27  \\ \hline
\multicolumn{2}{|l|}{MLP-Small} & 73.29 & \multicolumn2{l|}{MLP-Large}                          &   78.33 \\ \hline
\end{tabular}
\caption{Accuracy of slim KANs, MLP with comparable parameters (MLP-small), and MLP with an enlarged model size (MLP-large), when combined with a one-layer ResNet, for the image classification task on the CIFAR-10 dataset.}
\label{tab_cifar10}
\end{table}

\section{Conclusion}
In this paper, we introduce the  concept of low tensor-rank adaptation (LoTRA) for the transfer learning of KANs, inspired by Tucker decomposition in tensors and the success of LoRA for matrix parameter updates. 
We begin by empirically observing that both KAN parameters and fine-tuning updates exhibit a low tensor-rank structure, which motivates us to develop LoTRA as an efficient parameter update method.
We then theoretically establish the expressiveness of LoTRA based on Tucker decomposition approximations. Additionally, we propose a theoretically grounded learning rate selection strategy for efficient training of LoTRA, providing theoretical insights for practical implementation. Our analysis further reveals that applying identical learning rates to all LoTRA components is inefficient. 
Beyond theoretical insights, we explore the practical applications of LoTRA, particularly in fine-tuning KANs for solving PDEs and slimming KANs models. Experimental results validate our proposed learning rate selection strategy and demonstrate the effectiveness of LoTRA in fine-tuning KANs for solving PDEs. Furthermore, we evaluate Slim KANs in function representation and image classification tasks, showing that slim KANs maintain satisfying performance and significantly reduce the number of parameters. This is the first paper studying the transfer learning and fine-tuning of KANs with LoTRA.

Although our study mainly focuses on fine-tuning KANs using LoTRA for solving various PDEs, further exploration of LoTRA for broader transfer learning tasks remains an important direction. A deeper theoretical analysis of LoTRA is needed to enhance our understanding of its underlying properties, leading to better model interpretability and practical implementation. 
Future research could further refine its theoretical foundations and explore its practical integration into more complex deep learning architectures.

\newpage

\appendix

\section{Technical Proofs}

\subsection{Proof for Theorem 1}
\label{appendix_proof1}
For notational simplicity, we denote $\bz_{\ell,\text{ft}}$ and $\bz_{\ell,\text{tg}}$ as the $\ell$-th layer outputs of the fine-tuned model $\Psi_{\text{ft}}$ and the target model $\Psi_{\text{tg}}$, respectively. Then, we have
   \begin{equation*}
       \begin{split}
           & \quad \left\|\Psi_{\text{ft}}\left(\bm{x}\right) -  \Psi_{\text{tg}}\left(\bm{x}\right)\right\|_2 \\
           & = \left\| \Phi_{L}\left(\bz_{\ell,\text{ft}};  \mathcal{A}_{L,\text{ft}} \right) - \Phi_{L}\left(\bz_{\ell,\text{tg}};  \mathcal{A}_{L,\text{tg}} \right) \right\|_2 \\
           & \leq \left\| \Phi_{L}\left(\bz_{\ell,\text{ft}};  \mathcal{A}_{L,\text{ft}} \right) - \Phi_{L}\left(\bz_{\ell,\text{ft}};  \mathcal{A}_{L,\text{tg}} \right)\right\|_2\\
           & \quad + \left\| \Phi_{L}\left(\bz_{\ell,\text{ft}};  \mathcal{A}_{L,\text{tg}} \right) - \Phi_{L}\left(\bz_{\ell,\text{tg}};  \mathcal{A}_{L,\text{tg}} \right)\right\|_2.
       \end{split}
   \end{equation*} 
We define $\hat{\bz}_{\ell} \in \mathbb{R}^{n_{\ell} n_{d}}$ as $\hat{z}_{\ell,k+(p-1)n_d} = b_{k}\left(\phi\left(z_{\ell,p}\right)\right)$, for $k \in [n_d]$ and $p \in [n_{\ell}]$. Let $\bm{A}_{\ell}^{(2)}$ denote the mode-2 unfolding of the tensor $\mathcal{A}_{\ell}$, then we have
\begin{equation*}
    \begin{split}
        & \quad \left\| \Phi_{L}\left(\bz_{\ell,\text{ft}};  \mathcal{A}_{L,\text{ft}} \right) - \Phi_{L}\left(\bz_{\ell,\text{ft}};  \mathcal{A}_{L,\text{tg}} \right)\right\|_2 \\
        & = \left\| \left(\bA_{L,\text{ft}}^{(2)} - \bA_{L,\text{tg}}^{(2)} \right) \hat{\bz}_{L,\text{ft}}\right\|_2 \\
        & \leq \left\| \bA_{L,\text{ft}}^{(2)} - \bA_{L,\text{tg}}^{(2)}\right\|_{F} \cdot \left\| \hat{\bz}_{L,\text{ft}}\right\|_2 \\
        & = \left\| \mathcal{A}_{L,\text{ft}}^{(2)} - \mathcal{A}_{L,\text{tg}}^{(2)} \right\|_{F} \cdot \left\| \hat{\bz}_{L,\text{ft}}\right\|_2 \\
        & \leq B \sqrt{n_{L} n_{d}} \cdot \left( \sum_{r = r_{L,1} + 1}^{n_{L}} \sigma_{r} \left(\bm{E}_{L}^{(1)} \right)^2 + \sum_{r = r_{L,2} + 1}^{n_{L + 1}} \sigma_{r} \left(\bm{E}_{L}^{(2)} \right)^2 \right. \\
        & \quad \left. + \sum_{r = r_{L,3} + 1}^{n_{d}} \sigma_{r} \left(\bm{E}_{L}^{(3)} \right)^2\right)^{1/2},
    \end{split}
\end{equation*}
where $\left\| \hat{\bz}_{L,\text{ft}}\right\|_2$ is uniformly bounded by $B \sqrt{n_{L} n_{d}}$ under Assumption 1.

Furthermore,
\begin{equation*}
    \begin{split}
        & \quad \left\| \Phi_{L}\left(\bz_{\ell,\text{ft}};  \mathcal{A}_{L,\text{tg}} \right) - \Phi_{L}\left(\bz_{\ell,\text{tg}};  \mathcal{A}_{L,\text{tg}} \right)\right\|_2 \\
        & = \left\| \bA_{L,\text{tg}}^{(2)}\left(\hat{\bz}_{\ell,\text{ft}} - \hat{\bz}_{\ell,\text{tg}}\right)\right\|_2\\
        & \leq \left\| \bA_{L,\text{tg}}^{(2)} \right\|_{F} \cdot \left\| \hat{\bz}_{\ell,\text{ft}} - \hat{\bz}_{\ell,\text{tg}}\right\|_2 \\
        & = \left\| \mathcal{A}_{L,\text{tg}} \right\|_{F} \cdot \left\| \hat{\bz}_{\ell,\text{ft}} - \hat{\bz}_{\ell,\text{tg}}\right\|_2\\
        & \leq M \cdot L \sqrt{n_d} \left\| \bz_{\ell,\text{ft}} - \bz_{\ell,\text{tg}}\right\|_2.
    \end{split}
\end{equation*}
Applying induction from $\ell=1$ to $L$, we obtain the desired result.

\subsection{Derivation for $\Delta \Psi_{t,\text{ft}}$}
\label{appendix_derivation1}
\begin{equation*}
    \begin{split}
        & \quad \Delta \Psi_{t,\text{ft}} := \Psi_{t,\text{ft}} - \Psi_{t-1,\text{ft}} \\
        & = \sum_{p=1}^{r_1} \sum_{q=1}^{r_2} \sum_{k=1}^{r_3} g^{p,q,k}_{t} \cdot \left( \bm{u}^{(1), p \top}_{t} \bm{X} \bm{u}^{(3), k}_{t}\right) \cdot \bm{u}^{(2), q}_{t}\\
        & \quad - \sum_{p=1}^{r_1} \sum_{q=1}^{r_2} \sum_{k=1}^{r_3} g^{p,q,k}_{t-1} \cdot \left( \bm{u}^{(1), p \top}_{t-1} \bm{X} \bm{u}^{(3), k}_{t-1}\right) \cdot \bm{u}^{(2), q}_{t-1} \\
        & = \sum_{p=1}^{r_1} \sum_{q=1}^{r_2} \sum_{k=1}^{r_3} \left[g^{p,q,k}_{t-1} - \eta_{0} \cdot \left(\bm{v}_{t-1}^{\top} \bm{u}_{t-1}^{(2), q} \right) \cdot  \left( \bm{u}_{t-1}^{(1), p \top} \bm{X} \bm{u}_{t-1}^{(3),k}\right) \right] \cdot\\
        &  \cdot \left[ \bm{u}_{t-1}^{(1), p} - \eta_{1} \cdot \left( \sum_{q^{\prime}=1}^{r_2} \sum_{k^{\prime}=1}^{r_3} g_{t-1}^{p,q^{\prime},k^{\prime}} \cdot \left(\bm{v}_{t-1}^{\top} \bm{u}_{t-1}^{(2), q^{\prime}} \right) \cdot \bm{X} \bm{u}_{t-1}^{(3), k^{\prime}}\right) \right]^{\top}\\
        & \bm{X}\left[ \bm{u}_{t-1}^{(3), k} - \eta_{3} \cdot \left(\sum_{p^{\prime}=1}^{r_1} \sum_{q^{\prime}=1}^{r_2} g_{t-1}^{p^{\prime}, q^{\prime},k} \cdot \left(\bm{v}_{t-1}^{\top} \bm{u}_{t-1}^{(2), q^{\prime}} \right) \bm{X}^{\top} \bm{u}_{t-1}^{(1), p^{\prime}} \right)\right]\\
        & \cdot \left[ \bm{u}_{t-1}^{(2),k} - \eta_2 \cdot \left(\sum_{p^{\prime}=1}^{r_1} \sum_{k^{\prime}=1}^{r_3}g_{t-1}^{p^{\prime},q,k^{\prime}} \cdot \left( \bm{u}_{t-1}^{(1),p^{\prime} \top} \bm{X} \bm{u}_{t-1}^{(3), k^{\prime}}\right) \cdot \bm{v}_{t-1} \right)\right] \\
        & \quad - \sum_{p=1}^{r_1} \sum_{q=1}^{r_2} \sum_{k=1}^{r_3} g^{p,q,k}_{t-1} \cdot \left( \bm{u}^{(1), p \top}_{t-1} \bm{X} \bm{u}^{(3), k}_{t-1}\right) \cdot \bm{u}^{(2), q}_{t-1} \\
        & \approx \sum_{p=1}^{r_1} \sum_{q=1}^{r_2} \sum_{k=1}^{r_3} \delta_{t,0}^{p,q,k} + \sum_{p=1}^{r_1} \delta_{t,1}^{p} + \sum_{q=1}^{r_2} \delta_{t,2}^{q}  + \sum_{k=1}^{r_3} \delta_{t,3}^{k},
    \end{split}
\end{equation*}

\subsection{Proof for Theorem 2}
\label{appendix_proof2}
    If $r_1$, $r_2$, and $r_3$ are fixed constants and $n$ and $n_d$ are sufficiently large, then with high probability, the initialization satisfies
    \begin{equation}
    \label{eq_1}
        \begin{split}
            & \left|\bm{u}_{0}^{(1),p \top} \bm{X} \bm{u}_{0}^{(3),k}\right|   = \Theta \left(1 \right),\\
            & \left|\bm{u}_{0}^{(1),p \top} \bm{X} \bm{X}^{\top} \bm{u}_{0}^{(1),p} \right| = \Theta \left( n_d \right),\\
            & \left|\bm{u}_{0}^{(3),k \top} \bm{X} \bm{X}^{\top} \bm{u}_{0}^{(3),k} \right| = \Theta \left( n\right),\\
            & \left\| \bm{u}_{0}^{(2),q}\right\|_2  = \Theta \left( 1\right),\quad\bm{v}_{0}^{\top} \bu_{0}^{(2),q} = \Theta\left(1\right),
        \end{split}
    \end{equation}
    for $p \in [r_1]$, $q \in [r_2]$, and $k \in [r_3]$. 
    To achieve the efficient training of $g^{p,q,k}$ (i.e., $\delta_{t,0}^{p,q,k}$) in Definition 1, we require the condition to hold at $t=2$, equivalently, $\eta_{0} = \Theta\left( 1\right)$. Under this setting, we obtain $\left|g_{1}^{p,q,k}\right| = \Theta \left( 1\right)$.
    If all learning rates are set to the same order of magnitude, we have $\left\| \delta_{t,1}^{p}\right\|_2 = \Theta \left( n_d \right)$ and $\left\| \delta_{t,3}^{k}\right\|_2 = \Theta \left( n \right)$ (with $t=2$), which contradicts the conditions required for efficient training. 
    Furthermore, if we assume that relations in \Cref{eq_1} hold for $t \geq 0$, then we can similarly deduce that $\eta_{0} = \Theta(1)$, leading to the violation of conditions for efficient training.

    Now, we begin to verify that setting learning rates as $\eta_0 = \Theta\left(1\right)$, $\eta_1 = \Theta\left(n^{-1}\right)$, $\eta_2 = \Theta\left(1\right)$, and $\eta_3 = \Theta\left(n_d^{-1}\right)$, ensures the satisfaction of the conditions of efficient training.
    We first prove by induction that \Cref{eq_1} holds for all $t \geq 0$, i.e.,
    \begin{equation}
    \label{eq_2}
        \begin{split}
            & \left|\bm{u}_{t}^{(1),p \top} \bm{X} \bm{u}_{t}^{(3),k}\right|   = \Theta \left(1 \right),\\
            & \left|\bm{u}_{t}^{(1),p \top} \bm{X} \bm{X}^{\top} \bm{u}_{t}^{(1),p} \right| = \Theta \left( n_d \right),\\
            & \left|\bm{u}_{t}^{(3),k \top} \bm{X} \bm{X}^{\top} \bm{u}_{t}^{(3),k} \right| = \Theta \left( n\right),\\
            & \left\| \bm{u}_{t}^{(2),q}\right\|_2  = \Theta \left( 1\right),\quad\bm{v}_{t}^{\top} \bu_{t}^{(2),q} = \Theta\left(1\right),
        \end{split}
    \end{equation}
    for $p \in [r_1]$, $q \in [r_2]$, $k \in [r_3]$, and $t \geq 0$. Suppose that those relations in \Cref{eq_2} are satisfied at $t \geq 0$.
    First, we analyze the update of $g^{p,q,k}_{t+1}$ with
    \begin{equation*}
        \begin{split}
            & \quad \left| g^{p,q,k}_{t+1} - g^{p,q,k}_{t}\right|\\
            & = \eta_{0} \cdot \left|\bm{v}_{t}^{\top} \bm{u}_{t}^{(2), q} \right| \cdot  \left| \bm{u}_{t}^{(1), p \top} \bm{X} \bm{u}_{t}^{(3),k}\right| \\
            & = \Theta \left( 1\right).
        \end{split}
    \end{equation*}
    Therefore, we have $\left| g^{p,q,k}_{t+1} \right| = \Theta \left(1 \right)$, for $p \in [r_1]$, $q \in [r_2]$, and $k \in [r_3]$. 

For $ \bm{u}_{t+1}^{(1),p}$, we obtain  
    \begin{equation*}
        \begin{split}
            \left\| \Delta \bm{u}_{t+1}^{(1),p}\right\|_2 & = \left\| \bm{u}_{t+1}^{(1),p} - \bm{u}_{t}^{(1),p}\right\|_2 \\
            & \leq \eta_1 \cdot \sum_{q^{\prime}=1}^{r_2} \sum_{k^{\prime}=1}^{r_3} \left|g_{t}^{p,q^{\prime},k^{\prime}}\right| \cdot \left|\bm{v}_{t}^{\top} \bm{u}_{t}^{(2), q^{\prime}} \right| \cdot \left\|\bm{X} \bm{u}_{t}^{(3), k^{\prime}} \right\|_2\\
            & = \Theta\left( n^{-1/2}\right),
        \end{split}
    \end{equation*}
    then 
    \begin{equation*}
        \begin{split}
            & \quad \left|\bm{u}_{t+1}^{(1),p \top} \bm{X} \bm{X}^{\top} \bm{u}_{t+1}^{(1),p} - \bm{u}_{t}^{(1),p \top} \bm{X} \bm{X}^{\top} \bm{u}_{t}^{(1),p}\right|\\
            & \leq 2 \left\| \Delta \bm{u}_{t+1}^{(1),p}\right\|_2 \cdot \left\| \bm{X}\right\|_2 \cdot \sqrt{\left|\bm{u}_{t}^{(1),p \top} \bm{X} \bm{X}^{\top} \bm{u}_{t}^{(1),p}\right|}\\
            & \quad + \left\| \Delta \bm{u}_{t+1}^{(1),p}\right\|_2^2 \cdot \left\| \bm{X}\right\|_2^2 \\
            & \leq \Theta\left( n_d \right),
        \end{split}
    \end{equation*}
    which guarantees that $\left|\bm{u}_{t+1}^{(1),p \top} \bm{X} \bm{X}^{\top} \bm{u}_{t+1}^{(1),p}\right| = \Theta \left(n_d \right)$. 
    Similarly, we have
        \begin{equation*}
        \begin{split}
            & \quad \left\| \Delta \bm{u}_{t+1}^{(3),k}\right\|_2\\
            & = \left\| \bm{u}_{t+1}^{(3),k} - \bm{u}_{t}^{(3),k}\right\|_2 \\
            & = \eta_3 \cdot \sum_{p^{\prime}=1}^{r_1} \sum_{q^{\prime}=1}^{r_2} \left|g_{t}^{p^{\prime}, q^{\prime},k}\right| \cdot \left|\bm{v}_{t}^{\top} \bm{u}_{t}^{(2), q^{\prime}} \right| \cdot \left\| \bm{X}^{\top} \bm{u}_{t}^{(1), p^{\prime}}\right\|_2 \\
            & = \Theta\left( n_{d}^{-1/2}\right),
        \end{split}
    \end{equation*}
    and
            \begin{equation*}
        \begin{split}
            & \quad \left|\bm{u}_{t+1}^{(3),k \top} \bm{X}^{\top} \bm{X} \bm{u}_{t+1}^{(3),k} - \bm{u}_{t}^{(3),k \top} \bm{X}^{\top} \bm{X} \bm{u}_{t}^{(3),k}\right|\\
            & \leq 2 \left\| \Delta \bm{u}_{t+1}^{(3),k}\right\|_2 \cdot \left\| \bm{X}\right\|_2 \cdot \sqrt{\left|\bm{u}_{t}^{(3),k \top} \bm{X}^{\top} \bm{X} \bm{u}_{t}^{(3),k}\right|}\\
            & \quad + \left\| \Delta \bm{u}_{t+1}^{(3),k}\right\|_2^2 \cdot \left\| \bm{X}\right\|_2^2 \\
            & \leq \Theta\left( n_{\ell} \right),
        \end{split}
    \end{equation*}
    which ensures that $\left| \bm{u}_{t+1}^{(3),k \top} \bm{X}^{\top} \bm{X} \bm{u}_{t+1}^{(3),k} \right| = \Theta(n)$.
    Moreover, 
    \begin{equation*}
        \begin{split}
            & \quad \left| \bm{u}_{t+1}^{(1), p \top} \bm{X} \bm{u}_{t+1}^{(3),k} - \bm{u}_{t}^{(1), p \top} \bm{X} \bm{u}_{t}^{(3),k}\right| \\
            & \leq \left\| \Delta \bm{u}_{t+1}^{(1),p}\right\|_2 \cdot \sqrt{\left|\bm{u}_{t}^{(3),k \top} \bm{X}^{\top} \bm{X} \bm{u}_{t}^{(3),k}\right|} \\
            & \quad + \left\| \Delta \bm{u}_{t+1}^{(3),k}\right\|_2 \cdot \sqrt{\left|\bm{u}_{t}^{(1),p \top} \bm{X} \bm{X}^{\top} \bm{u}_{t}^{(1),p}\right|} \\
            & \quad + \left\| \Delta \bm{u}_{t+1}^{(1),p}\right\|_2 \cdot \left\| \Delta \bm{u}_{t+1}^{(3),k}\right\|_2 \cdot \left\| \bm{X} \right\|_2\\
            & = \Theta\left( 1\right),
        \end{split}
    \end{equation*}
    we have $\bm{u}_{t+1}^{(1), p \top} \bm{X} \bm{u}_{t+1}^{(3),k} = \Theta(1)$. Finally, we consider $\bm{u}_{t}^{(2)}$ with
    \begin{equation*}
        \begin{split}
            & \quad \left\| \Delta \bm{u}_{t+1}^{(2),q}\right\|_2 \\
            & = \left\| \bm{u}_{t+1}^{(2),q} - \bm{u}_{t}^{(2),q}\right\|_2 \\
            & = \eta_2 \cdot \sum_{p^{\prime}=1}^{r_1} \sum_{k^{\prime}=1}^{r_3} \left|g_{t-1}^{p^{\prime},q,k^{\prime}} \right| \cdot \left| \bm{u}_{t-1}^{(1),p^{\prime} \top} \bm{X} \bm{u}_{t-1}^{(3), k^{\prime}}\right| \cdot \left\|\bm{v}_{t-1} \right\|_2\\
            & = \Theta\left( 1\right),
        \end{split}
    \end{equation*}
    which implies $\left\| \bm{u}_{t+1}^{(2),q}\right\|_2 = \Theta(1)$. 
    Note that
    \begin{equation*}
        \Delta \Psi_{t+1,\text{ft}} = \sum_{q=1}^{r_2} a_{t,q} \bm{u}_{t}^{(2),q} + c_{t} \bm{v}_{t},
    \end{equation*}
    with some numbers $a_{t,q} = \Theta\left(1\right)$ and $c_t = \Theta\left(1\right)$, then we have
    \begin{equation*}
        \begin{split}
            \bm{v}_{t+1}^{\top} \bm{u}_{t+1}^{(2),q}
            & = \left( \bm{v}_{t} + \Delta \Psi_{t+1,\text{ft}} \right)^{\top} \left(\bm{u}_{t}^{(2),q}  + \Delta \bm{u}_{t+1}^{(2),q}  \right) \\
            & = \bm{v}_{t}^{\top} \bm{u}_{t}^{(2),q} + \Theta(1)\\
            & = \Theta\left(1\right).
        \end{split}
    \end{equation*}
    Therefore, all conditions in \Cref{eq_2} hold for $t+1$ as well. By induction, it follows that these conditions are satisfied for all $t$.

    Under the conditions in \Cref{eq_2} and the learning rates setup, where $\eta_0 = \Theta\left(1\right)$, $\eta_1 = \Theta\left(n^{-1}\right)$, $\eta_2 = \Theta\left(1\right)$, and $\eta_3 = \Theta\left(n_d^{-1}\right)$, we can verify that the quantities $\left\{\left\|\delta_{t,0}^{p,q,k}\right\|_2 \right\}$, $\left\{\left\|\delta_{t,1}^{p}\right\|_2 \right\}$, $\left\{\left\|\delta_{t,2}^{q}\right\|_2 \right\}$, and $\left\{\left\|\delta_{t,3}^{k}\right\|_2 \right\}$ are all of order $ \Theta\left(1\right)$ with respect to the model size $(n,m,n_d)$, for all $i \in \{0,1,2,3\}$, $p \in [r_1]$, $q \in [r_2]$, $k \in [r_3]$, and $t \geq 2$.

\newpage

\bibliographystyle{IEEEtran}
\bibliography{main}

\newpage

\end{document}